\documentclass[letterpaper]{article} % DO NOT CHANGE THIS
\usepackage{aaai2026}  % DO NOT CHANGE THIS
\usepackage{times}  % DO NOT CHANGE THIS
\usepackage{helvet}  % DO NOT CHANGE THIS
\usepackage{courier}  % DO NOT CHANGE THIS
\usepackage[hyphens]{url}  % DO NOT CHANGE THIS
\usepackage{graphicx} % DO NOT CHANGE THIS
\usepackage{natbib}  % DO NOT CHANGE THIS AND DO NOT ADD ANY OPTIONS TO IT
\usepackage{caption} % DO NOT CHANGE THIS AND DO NOT ADD ANY OPTIONS TO IT
\usepackage{algorithm}
\usepackage{algorithmic}
\usepackage[most]{tcolorbox}
\usepackage{graphicx}
\usepackage{subcaption}
\usepackage{booktabs}
\usepackage{newfloat}
\usepackage{listings}
\newcommand{\firstauthormark}{\textsuperscript{†}}
\newcommand{\corresauthormark}{\textsuperscript{*}}
\DeclareCaptionStyle{ruled}{labelfont=normalfont,labelsep=colon,strut=off} % DO NOT CHANGE THIS
\floatstyle{ruled}
\newfloat{listing}{tb}{lst}{}
\floatname{listing}{Listing}
\title{\textbf{MedAtlas}: Evaluating LLMs for Multi-Round, Multi-Task Medical Reasoning Across Diverse Imaging Modalities and Clinical Text}
\author{
    Ronghao Xu\firstauthormark\textsuperscript{\rm 1,2},
    Zhen Huang\firstauthormark\textsuperscript{\rm 3,4},
    Yangbo Wei\textsuperscript{\rm 4,5},
    Xiaoqian Zhou\textsuperscript{\rm 1,2},
    Zikang Xu\textsuperscript{\rm 6},
    Ting Liu\textsuperscript{\rm 1,2},
    Zihang Jiang\corresauthormark\textsuperscript{\rm 1,2},
    S. Kevin Zhou\corresauthormark\textsuperscript{\rm 1,2}
}
\affiliations{
    \textsuperscript{\rm 1}School of Biomedical Engineering, Division of Life Sciences and Medicine,\\
    University of Science and Technology of China,
    Hefei, Anhui 230026, P.R. China\\
    
    \textsuperscript{\rm 2}Suzhou Institute for Advanced Research,\\
    University of Science and Technology of China,
    Suzhou, Jiangsu 215123, P.R. China\\
    
    \textsuperscript{\rm 3}School of Computer Science and Technology,
    University of Science and Technology of China,
    Hefei 230026, P.R. China\\
    
    \textsuperscript{\rm 4}School of Information Science and Technology,
    Eastern Institute of Technology,
    Ningbo 315200, P.R. China\\
    
    \textsuperscript{\rm 5}Shanghai Jiao Tong University,
    Shanghai 200030, P.R. China\\
    
    \textsuperscript{\rm 6}Anhui Province Key Laboratory of Biomedical Imaging and Intelligent Processing,\\
    Institute of Artificial Intelligence,
    Hefei Comprehensive National Science Center,
    Hefei 230026, P.R. China\\
    
    \textsuperscript{†}These authors contributed equally\\
    \textsuperscript{*}Corresponding authors: 
    jzh0103@ustc.edu.cn,
    skevinzhou@ustc.edu.cn
}

\usepackage{bibentry}
\begin{document}

\maketitle

\begin{abstract}
Artificial intelligence has demonstrated significant potential in clinical decision-making; however, developing models capable of adapting to diverse real-world scenarios and performing complex diagnostic reasoning remains a major challenge. Existing medical multi-modal benchmarks are typically limited to single-image, single-turn tasks, lacking multi-modal medical image integration and failing to capture the longitudinal and multi-modal interactive nature inherent to clinical practice. To address this gap, we introduce \textbf{MedAtlas}, a novel benchmark framework designed to evaluate large language models on realistic medical reasoning tasks. \textbf{MedAtlas} is characterized by four key features: multi-turn dialogue, multi-modal medical image interaction, multi-task integration, and high clinical fidelity. It supports four core tasks: open-ended multi-turn question answering, closed-ended multi-turn question answering, multi-image joint reasoning, and comprehensive disease diagnosis. Each case is derived from real diagnostic workflows and incorporates temporal interactions between textual medical histories and multiple imaging modalities, including CT, MRI, PET, ultrasound, and X-ray, requiring models to perform deep integrative reasoning across images and clinical texts. \textbf{MedAtlas} provides expert-annotated gold standards for all tasks. Furthermore, we propose two novel evaluation metrics: Round Chain Accuracy and Error Propagation Resistance. Benchmark results with existing multi-modal models reveal substantial performance gaps in multi-stage clinical reasoning. \textbf{MedAtlas} establishes a challenging evaluation platform to advance the development of robust and trustworthy medical AI.
\end{abstract}


\section{Introduction}

Clinical multimodal reasoning is a core challenge in intelligent healthcare, requiring AI models to simultaneously process imaging examinations, clinical history, and progressive inquiry information to support diagnostic decision-making. In real-world clinical scenarios, physicians often reach a final diagnosis through multiple rounds of inquiry and various imaging examinations: alternately reviewing multimodal medical images (e.g., CT, MRI, ultrasound), probing key clinical manifestations, and integrating longitudinal examination results to form a comprehensive judgment. 

Existing medical AI benchmarks are mostly limited to single-round, single-image visual question answering (VQA) settings. This limitation prevents models from addressing complex diagnostic needs in real-world scenarios, such as differential diagnosis based on multi-image comparison (``Compared to three months ago, the enlargement of this lesion suggests what possibility?'') or tasks requiring stepwise reasoning (``Based on the current MRI findings, should the next step be a biopsy or a PET-CT?''). The absence of such benchmarks constrains the development of practical medical assistants.

The limitations of existing benchmarks are reflected in five dimensions:  
(1) \textbf{Task simplicity}: VQA datasets (e.g., VQA-RAD) only support single-round question answering, while classification datasets (e.g., CheXpert) lack multi-round, free-form image-text interaction.  
(2) \textbf{Modality isolation}: Most benchmarks do not require cross-modal association (e.g., simultaneously parsing multi-modal images and textual information).  
(3) \textbf{Reasoning discontinuity}: Even if multi-task datasets exist (e.g., PMC-VQA includes question answering and report generation tasks), they fail to construct coherent chains of clinical reasoning. Such fragmented evaluation deviates significantly from real diagnostic workflows—radiologists need to compare patients' historical images (multi-image), incorporate pathology reports (multi-text), and perform multi-round inquiry (multi-turn) for stepwise diagnosis, capabilities that remain untested in current benchmarks.  
(4) \textbf{Multi-image integration}: There is a lack of evaluation for the integrated understanding and judgment of medical images from various modalities (X-ray, MRI, CT, PET, etc.).

\begin{figure*}[t]
    \centering
    \includegraphics[width=0.8\textwidth]{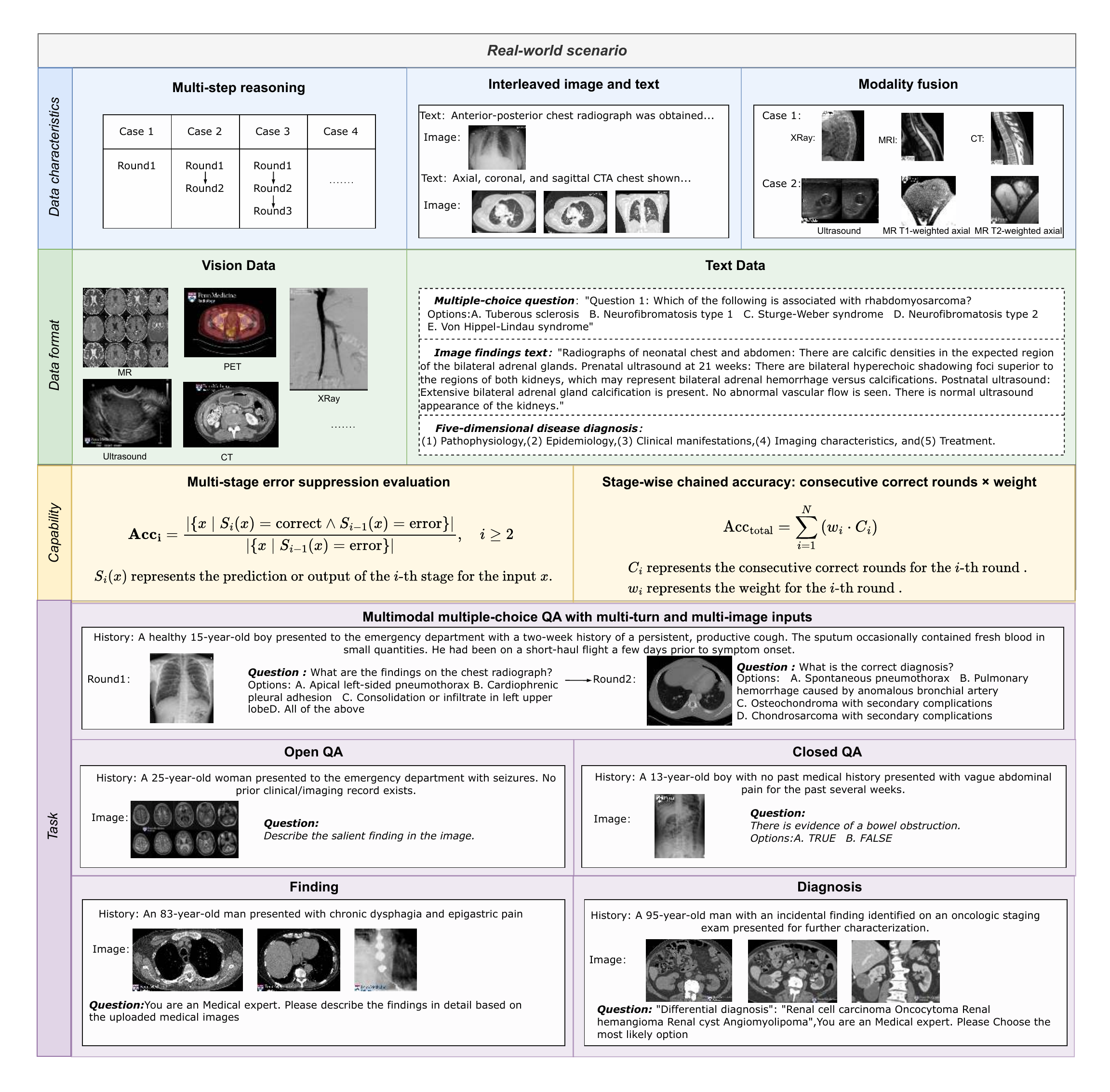}
    \caption{\textbf{Overview of the \textbf{MedAtlas} Benchmark.} \textbf{MedAtlas} captures real-world diagnostic workflows via multi-round, multi-modal, and multi-image reasoning. It supports diverse tasks including open-ended QA, closed QA, multi-image reasoning, and multi-modal multi-choice QA. The benchmark features rich clinical inputs (images + text), realistic case structures, and two specialized metrics: Stage Chain Accuracy and Error Propagation Suppression.}
    \label{fig:overview}
\end{figure*}
This work introduces \textit{\textbf{MedAtlas}}, a benchmark dataset specifically designed to address the identified challenges. The dataset is constructed from real clinical cases collected from publicly available sources and explicitly captures the core characteristics of diagnostic reasoning involving multi-turn, multimodal, and multi-image contexts. 

Each case is organized as a sequence of question-answer (Q\&A) pairs spanning multiple diagnostic stages. It begins with the clinical history (a textual description of the patient's background), followed by several imaging stages. The first stage presents an initial set of images (e.g., radiographs) along with corresponding questions; the second stage introduces advanced imaging (e.g., contrast-enhanced MRI/CT) and follow-up questions that depend on information from the previous stage; if necessary, the sequence extends to a third stage (e.g., specific-position MRI accompanied by new queries), thereby simulating the progressive process of clinical examination and inquiry. 

All Q\&A pairs collectively form a reasoning chain: subsequent questions build upon earlier findings, and the narrative of each case unfolds step by step, culminating in a final diagnostic or therapeutic question. The dataset encompasses multiple imaging modalities, including X-ray, CT, MRI, ultrasound, and PET (with some cases containing mixed modalities). Many stages require the simultaneous interpretation of multiple images (e.g., multi-plane MRI or multi-view CT), thereby necessitating cross-image integration capabilities.  

Additionally, each case is annotated with structured information: imaging findings (textual descriptions of key observations), diagnostic outcomes, and detailed disease information (including pathophysiology, epidemiology, clinical manifestations, imaging characteristics, and treatment plans). Such annotations enable extended evaluations of models on tasks such as generating comprehensive clinical summaries and recalling domain-specific knowledge. 

By systematically addressing the five major limitations of existing datasets, \textit{\textbf{MedAtlas}} provides a benchmark that is more closely aligned with real-world clinical scenarios than current alternatives.

\textbf{Main Contributions:}  
\begin{itemize}
\item \textbf{A novel benchmark dataset:} The first multi-task medical benchmark featuring multi-turn dialogue, multi-modal multi-image inputs, and explicitly defined clinical reasoning chains. \textbf{MedAtlas} consists of a diverse collection of cases covering a wide range of conditions. Each case is richly annotated with patient history, multiple imaging studies (Rounds), sequential QA pairs, and additional context like radiological findings and disease background. This dataset aims to bridge the gap between simplified VQA tasks and the complexities of actual clinical diagnosis.
\item \textbf{Innovative evaluation metrics:} In addition to conventional QA accuracy, two new metrics are introduced: \textit{Stage-Chain Accuracy}, which measures the longest sequence of consecutive correct diagnostic stages, and the \textit{Error Propagation Suppression Coefficient}, which evaluates a model's ability to maintain downstream performance despite upstream errors.  
\item \textbf{Comprehensive evaluation of state-of-the-art vision-language models:} A systematic assessment of the latest 2025 versions of prominent multi-turn, multi-image capable large models, including GPT-4o, Qwen, Claude, and LLaVA.  
\end{itemize}

By providing the \textbf{MedAtlas} dataset, an accompanying evaluation framework, and baseline results, this work aims to drive medical VQA models toward more complex, interpretable, and robust clinical reasoning. The benchmark highlights the limitations of current models in handling complex real-world scenarios and provides clear directions for future research.

\section{Related Work}
\label{sec:related_work}

Our work builds on prior research in Med-VQA, vision-language models (VLMs), multi-turn dialogue, and evaluation. Early Med-VQA datasets such as VQA-RAD~\cite{lau2022vqa_rad}, SLAKE~\cite{liu2021slake}, and PathVQA~\cite{he2020pathvqa} introduced radiology and pathology QA but were limited in scale, modality, and lacked multi-turn or multi-image reasoning. Larger datasets like PMC-VQA~\cite{zhang2023pmcvqa}, OmniMedVQA~\cite{liu2024omnimedvqa}, and MIMIC-Diff-VQA~\cite{hu2023mimic_diff_vqa} improved coverage and complexity, yet still mostly featured single-turn QA. Some works like RadVisDial explored synthetic multi-turn dialogue but remained narrow in scope. Parallel progress in general VLMs (e.g., GPT-4~\cite{openai2023gpt4}, Flamingo, BLIP-2~\cite{li2023blip2}, InstructBLIP~\cite{dai2023instructblip}, LLaVA~\cite{li2023llava}) and medical-specific variants (e.g., Med-Flamingo~\cite{moor2023medflamingo}, LLaVA-Med, RadFM~\cite{zhang2023radfm}) has enabled stronger multimodal reasoning. For evaluation, benchmarks like ConvBench~\cite{liu2024convbench}, SparklesEval~\cite{huang2023sparkles}, Mantis-Eval~\cite{jiang2024mantiseval}, and MMDU~\cite{chen2024mmdu} extend to multi-turn and multi-image contexts but remain largely non-medical. Our work addresses these gaps by integrating multi-turn dialogue and multi-image reasoning within a medical setting.

\begin{figure*}[t]
    \centering
    \includegraphics[width=\linewidth,trim=0cm 1cm 0cm 0cm, clip]{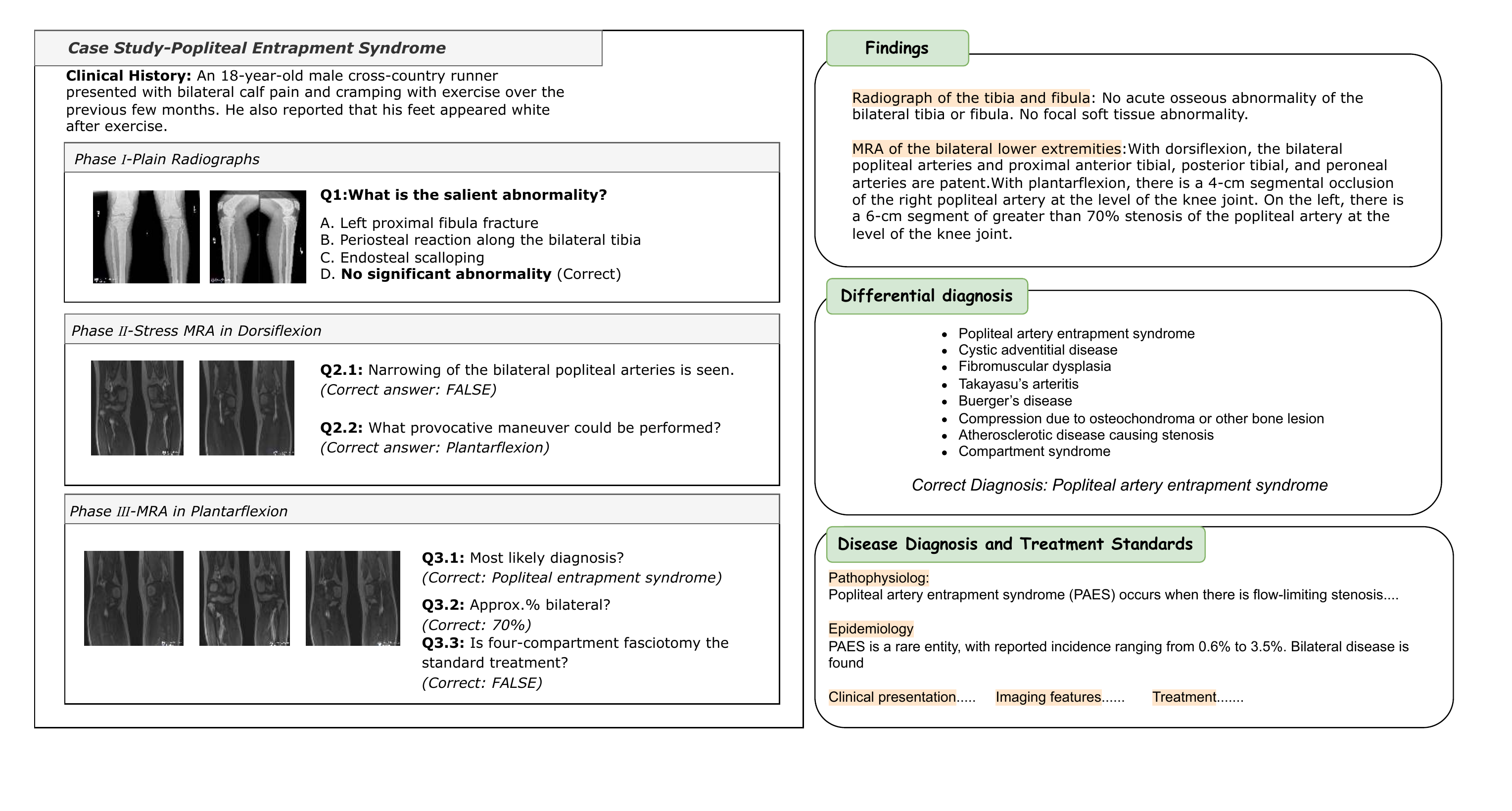}
    \caption{An example of a dataset includes multiple rounds of conversations and multi-image understanding.}
    \label{fig:example}
\end{figure*}

\section{\textbf{MedAtlas} Benchmark}
\label{sec:dataset}
% \begin{figure}[htbp]
%     \centering
%     \includegraphics[width=\textwidth]{net.png}
%     \caption{\textbf{Overview of the \textbf{MedAtlas} Benchmark.} The top row illustrates real‐world data characteristics captured by \textbf{MedAtlas}: (i) \emph{Multi‐step reasoning} chains spanning multiple diagnostic rounds, (ii) \emph{Interleaved image and text} inputs combining clinical history, findings descriptions, and multiple imaging modalities, and (iii) \emph{Modality fusion} scenarios where cases progress across X‐ray, ultrasound, CT, MRI, PET, etc. The middle row shows the dataset format: diverse \emph{vision data} (e.g., MR, PET, X‐ray, ultrasound, CT) paired with \emph{text data} including multiple‐choice questions, structured image findings, and five‐dimensional disease annotations. The third row presents the four primary tasks—Task 01: Open‐ended QA, Task 02: Multi‐image reasoning, Task 03: Closed QA, and Task 04: multi-modal multi‐choice QA with multi‐turn, multi‐image inputs—highlighting the progressive complexity of clinical interaction. The bottom row summarizes the core evaluation capabilities: \emph{Error Propagation Suppression} (quantifying recovery from early mistakes) and \emph{Stage‐wise Chained Accuracy} (measuring consecutive correct rounds weighted by clinical significance). Together, these components define \textbf{MedAtlas} as a comprehensive benchmark for multi‐round, multi‐modal, and multi‐image diagnostic reasoning.}
%     \label{fig:overview}
% \end{figure}

\textbf{MedAtlas} is designed to rigorously evaluate the ability of VLMs to engage in multi-round, multi-modal, and multi-image reasoning within the context of realistic clinical diagnostic workflows. It moves beyond static, single-instance VQA towards simulating the dynamic progression of patient cases.

\begin{figure*}[t]
    \centering
    \includegraphics[width=\linewidth, trim=0cm 0cm 0cm 0cm, clip]{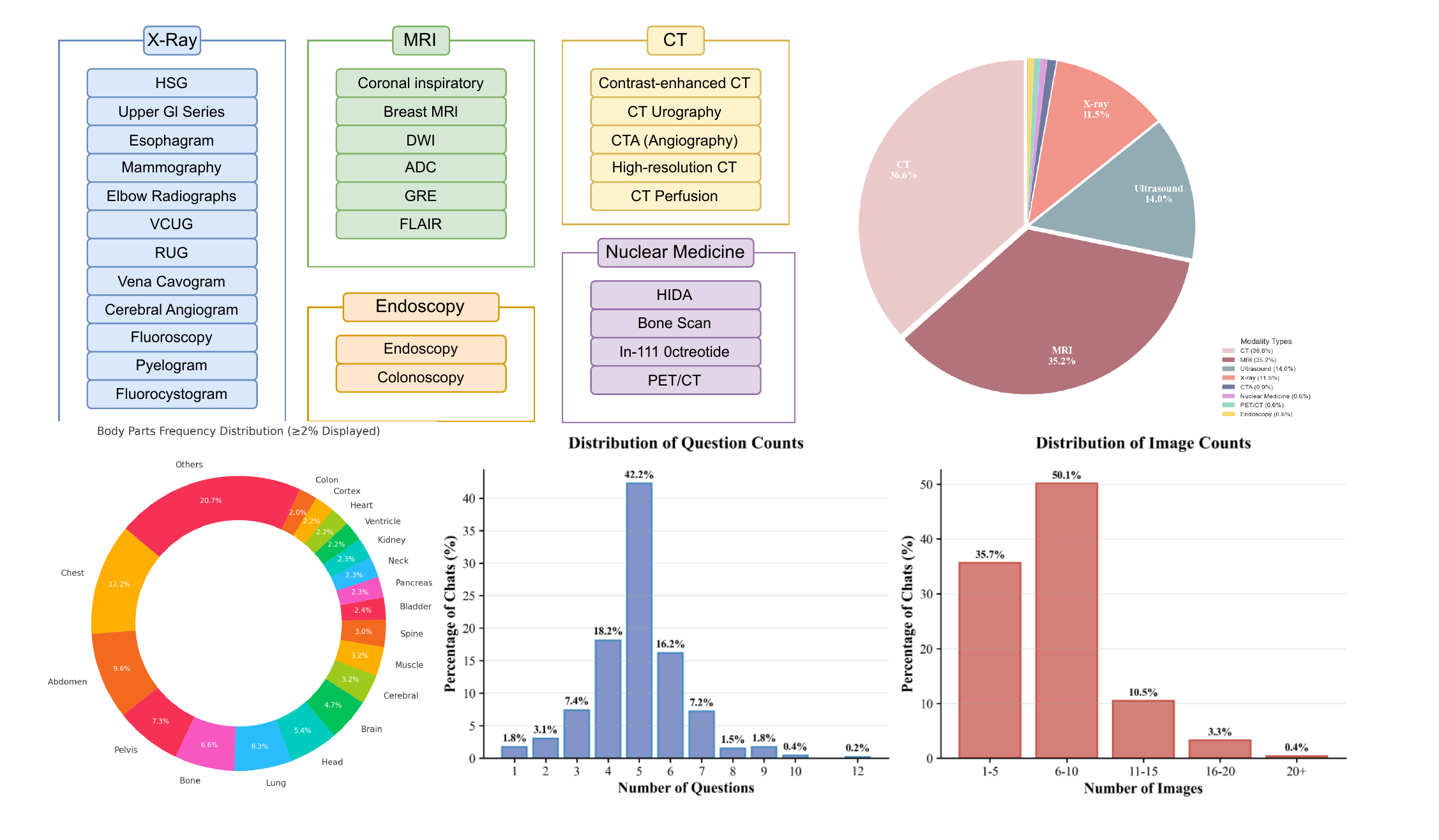}
    \caption{Medical Imaging Data Distribution and Analysis: This figure illustrates various imaging modalities (X-ray, MRI, CT, Nuclear Medicine, Ultrasound, Endoscopy) along with their specific procedures, modality usage proportions, body part frequency distribution, and the distributions of question and image counts. The data indicate that CT (36.4\%) and MRI (35.2\%) are the predominant imaging modalities, with chest and abdomen being the most frequently examined regions. Most conversations involve 3–6 questions and include 1–10 images.}
    \label{fig:image}
\end{figure*}

\section{Dataset Statistic}

In this section, we provide a comprehensive statistical overview of the multi-round, multi-modal dialogue reasoning task within our proposed benchmark, \textbf{MedAtlas} as shown in Figure~\ref{fig:overview}. This task is specifically designed for evaluating the capabilities of Visual Language Models (VLMs) in realistic clinical diagnostic scenarios involving sequential reasoning across multiple dialogue rounds, imaging modalities, and images.

\subsection{Overall Dataset Statistics}
As shown in Figure~\ref{fig:image},the multi-round, multi-modal task in \textbf{MedAtlas} comprises 804 clinically realistic patient cases, encompassing a total of 5632 medical images. Each case is structured into multi-turn dialogues with clinical reasoning tasks, containing a total of 1516 dialogue rounds and 4015 question-answer pairs. On average, each case includes approximately 1.94 dialogue rounds, with a median of 3 rounds per case; complex scenarios can extend up to about 10 rounds.

\subsection{Question and Answer Characteristics}
The questions in this task are predominantly structured as multiple-choice, with single-choice questions constituting the majority (1641 questions, accounting for 72.29\%), complemented by True/False questions (629 questions, 27.71\%). The average number of questions per dialogue round is approximately 4.96, mirroring realistic clinical diagnostic workflows.

The dataset features multiple-choice questions with varied numbers of options (ranging from 2 to 17), reflecting the diverse complexity of clinical scenarios. A detailed distribution is illustrated separately in our figures, highlighting the predominance of questions with 4 or fewer options, typical of clinical diagnostic and educational scenarios. Additionally, cases with higher numbers of options represent more complex scenarios that require extensive differential diagnosis and advanced clinical reasoning.

\subsection{Imaging Modalities}
The multi-round dialogue task within \textbf{MedAtlas} incorporates a wide range of medical imaging modalities. MRI (36.2\%) and X-ray (31.7\%) constitute the majority of the images, followed by CT (17.1\%) and ultrasound (8.6\%). Other modalities, including PET/CT, nuclear medicine, endoscopy, mammography, and angiography, are included in smaller proportions. Specific imaging techniques within each modality are detailed explicitly in accompanying figures. The extensive modality coverage in this task challenges VLMs to develop robust and modality-invariant medical image interpretation and reasoning skills.

\subsection{Question and Image Count Distribution}
The distribution of question counts per case in this task shows most cases (approximately 76.6\%) contain between 4 and 6 questions, with 42.2\% of cases containing exactly 5 questions. Only a minor proportion of cases involve fewer than 3 or more than 8 questions, ensuring both clinical relevance and complexity variability.

Regarding image count per case, most cases (50.1\%) have between 6 and 10 images, followed by 36.7\% with 1 to 5 images. Approximately 10.5\% of cases are more complex with 11–15 images, and only a minimal fraction (0.4\%) exceed 20 images. This structured diversity ensures comprehensive coverage of both routine and complex clinical scenarios.

\subsection{Clinical Reasoning and Complexity}
Questions in the multi-round, multi-modal dialogue task as shown in Figure \ref{fig:example} within \textbf{MedAtlas} follow a clear clinical reasoning progression, starting from initial observations, advancing through analysis and differential diagnosis, and culminating in final diagnostic conclusions and treatment recommendations. The dataset inherently facilitates multi-hop reasoning, as later questions frequently depend on accurate interpretation of previous images and findings. This design closely mirrors authentic clinical reasoning processes.

Overall, this task within \textbf{MedAtlas} represents a rigorous and clinically realistic benchmark. Its unique multi-round dialogue structure, diverse multi-modal inputs, and requirement for integrating multi-image information provide a comprehensive evaluation platform for testing VLMs' capabilities in clinical reasoning, medical knowledge retrieval, and diagnostic decision-making.

\section{Evaluation Metrics}

As discussed earlier, evaluating a model on \textbf{MedAtlas} requires both classical metrics and new, specialized metrics to capture reasoning chains. We detail our metrics and how they are computed.

\subsection{Standard VQA Metrics}
For each question in \textbf{MedAtlas}, if a standard short answer or multiple-choice answer is available, binary correctness is computed. The overall accuracy is defined as the proportion of fully correct answers; specifically, multiple-choice questions require selecting all correct options, and single-word or short-phrase responses must match the expected standardized strings. This approach aligns with existing VQA evaluation methodologies~\cite{lau2022vqa_rad,liu2021slake}. For open-ended questions requiring sentence-level answers, GPT-4o is employed to evaluate the semantic consistency between the generated answers and the reference answers, mitigating erroneous judgments caused by differences in phrasing or synonymous expressions.

\subsubsection{Stage Chain Accuracy (SCA).}
SCA measures how far a model can progress through a multi-Round task without error, emphasizing sequential consistency. 
For each case $c$, we identify the longest prefix of Rounds answered completely correctly. Let $L(c)$ denote this chain length, and let $\{w_1, w_2, \dots, w_K\}$ be Round weights satisfying $w_1 < w_2 < \dots < w_K$ to emphasize later Rounds. The per-case SCA score is then:
\begin{equation}
    \mathrm{SCA}(c) = w_{L(c)}.
\end{equation}

The overall SCA is computed as the average across all cases:
\begin{equation}
    \mathrm{SCA} = \frac{1}{|\mathcal{C}|} \sum_{c \in \mathcal{C}} \mathrm{SCA}(c),
\end{equation}
where $\mathcal{C}$ is the set of all cases. We also report the distribution of chain lengths (e.g., proportion of cases reaching Round I, Round II, etc.) to characterize where models tend to fail.

\paragraph{Interpretation.} 
A high SCA indicates that the model can sustain correct answers over multiple sequential Rounds, whereas a steep drop in chain length distribution highlights systematic failures in later Rounds. Unlike per-question accuracy, SCA reflects both accuracy and error ordering, thus directly measuring sequential reasoning consistency.

\subsection{Error Propagation Suppression Coefficient (EPSC)}
To quantify the degree to which early-stage errors affect downstream performance, we compute the \textit{Error Propagation Suppression Coefficient (EPSC)} across all multi-Round cases. We focus on error propagation from Round I to Round II, which is the most critical and widely available Round pair in our dataset.

\paragraph{Definition.} 
Let $A(\text{correct\_prev})$ denote the Round II accuracy across all cases where the model answered all Round I questions correctly. Similarly, let $A(\text{wrong\_prev})$ denote the Round II accuracy across all cases where the model made at least one error in Round I. The EPSC is then defined as:
\begin{equation}
    \mathrm{EPSC} = \frac{A(\text{wrong\_prev})}{A(\text{correct\_prev})}.
\end{equation}

Here,
\begin{equation}
    A(\text{correct\_prev}) = \frac{1}{|\mathcal{C}_{\text{correct}}|} \sum_{c \in \mathcal{C}_{\text{correct}}} 
    \frac{\sum_{q \in \mathcal{P}_2(c)} \mathbf{1}\left[ \hat{y}_q = y_q \right]}{|\mathcal{P}_2(c)|},
\end{equation}
\begin{equation}
    A(\text{wrong\_prev}) = \frac{1}{|\mathcal{C}_{\text{wrong}}|} \sum_{c \in \mathcal{C}_{\text{wrong}}} 
    \frac{\sum_{q \in \mathcal{P}_2(c)} \mathbf{1}\left[ \hat{y}_q = y_q \right]}{|\mathcal{P}_2(c)|},
\end{equation}
where $\mathbf{1}[\cdot]$ is an indicator function that equals $1$ if the condition is true and $0$ otherwise, 
$\mathcal{C}_{\text{correct}}$ is the set of cases where Round I is fully correct, 
$\mathcal{C}_{\text{wrong}}$ is the set of cases where Round I contains any incorrect answer, 
$\mathcal{P}_2(c)$ denotes all questions in Round II of case $c$, $\hat{y}_q$ is the model’s prediction, and $y_q$ is the ground truth.

\paragraph{Interpretation.}
An $\mathrm{EPSC}$ close to $1.0$ indicates that Round II accuracy is largely unaffected by errors in Round I, implying strong error suppression and robustness to upstream mistakes. Conversely, a low EPSC (e.g., $<0.7$) indicates substantial error propagation: when the model errs in Round I, its subsequent reasoning degrades significantly.

\paragraph{Implementation Detail.}
Unlike prior definitions restricted to per-case conditional analysis, we compute $\mathrm{EPSC}$ \textit{globally} across all cases in the dataset. This approach aggregates evidence from all available samples without requiring both ``correct'' and ``wrong'' Round I outcomes to appear within the same case, thereby yielding a more statistically stable estimate:
\begin{equation}
    \mathcal{C}_{\text{correct}} \cap \mathcal{C}_{\text{wrong}} = \emptyset.
\end{equation}
This global formulation better reflects overall model robustness to early-stage errors in multi-Round reasoning tasks.

\subsection{Findings Evaluation}

For the \textbf{Findings generation} task, the evaluation metrics include BLEU~\cite{papineni2002bleu} (measuring n-gram overlap), ROUGE-L~\cite{lin2004rouge} (assessing longest common subsequence overlap with reference text), and METEOR~\cite{banerjee2005meteor} (which accounts for synonyms via WordNet). Additionally, BERTScore is computed using \textbf{BioClinicalBERT}~\cite{lee2020biobert}, a domain-specific language model pretrained on biomedical and clinical corpora, to better capture semantic similarity and handle paraphrasing in clinical contexts. 

These metrics collectively quantify language fidelity relative to reference findings while incorporating domain-specific semantic understanding. For diagnostic evaluation, the task is treated as a classification problem: if the reference diagnosis is “torn ACL” and the model predicts “ACL tear,” it is considered correct after lemmatization. To ensure fairness, a custom normalization procedure is applied to medical terminology, accounting for variations such as singular vs. plural forms, minor spelling differences, and abbreviations versus full terms.

% \subsection{Aspect-based Knowledge}

% For \textit{Aspect-based Knowledge}, since each aspect is usually a few sentences, we again use BLEU/ROUGE for overlap~\cite{papineni2002bleu,lin2004rouge}, and a scoring rubric: for each aspect, we have a checklist of key points that should be mentioned (drawn from standard textbooks or the reference). We award points for each point covered and aggregate these to an accuracy percentage for each aspect and average across the five aspects. This yields the \textbf{Knowledge Aspect Accuracy}. We found that this granular evaluation is insightful -- models might know treatment guidelines well (scoring high on Treatment) but falter on epidemiology (which often involves numeric data or specific at-risk populations).
\subsection{Knowledge Accuracy (Knowledge Acc.) Calculation}
To better capture the composite capability of visual-language models (VLMs) across multiple dimensions, we introduce \textbf{Knowledge Accuracy (Knowledge Acc.)}, which aggregates four core metrics: Open QA Accuracy, Closed QA Accuracy, Multi-Image Findings, and Diagnosis Accuracy. 

\paragraph{Step 1: Metric Aggregation.}
We first compute the mean performance $\mu_i$ of model $i$ across these four metrics:
\begin{equation}
    \mu_i = \frac{\text{OpenQA}_i + \text{ClosedQA}_i + \text{MultiImage}_i + \text{Diagnosis}_i}{4}.
\end{equation}

\paragraph{Step 2: Z-score Standardization.}
To normalize and compare models on a common scale, we apply Z-score standardization:
\begin{equation}
    Z_i = \frac{\mu_i - \overline{\mu}}{\sigma},
\end{equation}
where $\overline{\mu}$ and $\sigma$ represent the mean and standard deviation of $\mu_i$ across all models.

\paragraph{Step 3: Logistic Mapping.}
Finally, to emphasize performance differences between high-performing and low-performing models, we map $Z_i$ into the $[0,1]$ range using a logistic transformation:
\begin{equation}
    \text{KnowledgeAcc}_i = \frac{1}{1 + e^{-k \cdot Z_i}},
\end{equation}
where $k$ is a scaling hyperparameter (we set $k=3$ in our experiments) controlling the steepness of the logistic curve.

\begin{table*} 
\centering
\caption{Normalized performance of general-purpose and medical-specific VLMs on \textbf{MedAtlas}, sorted by Knowledge Acc. (ascending).}
\label{tab:scaled-baseline-results}
\resizebox{\textwidth}{!}{
\begin{tabular}{lccccccc}
\toprule
Model & Open QA Acc. & Closed QA Acc. & Multi-Image Findings & Diagnosis & Knowledge Acc. & SCA & EPSC \\
\midrule
InternVL3-78B ~\cite{internvl3_2025}     & 10.80 & 25.21 & 32.03 & 6.50  & 0.01 & 0.25 & 0.16 \\
Qwen2.5-VL-7B  ~\cite{qwen2024}    & 25.21 & 14.80 & 32.03 & 6.50  & 0.01 & 0.30 & 0.20 \\
Qwen2.5-VL-32B ~\cite{qwen2024}    & 20.00 & 57.70 & 27.85 & 7.40  & 0.09 & 0.60 & 0.35 \\
Kimi-latest  ~\cite{kimi2025}      & 15.30 & 63.61 & 33.30 & 8.10  & 0.15 & 1.94 & 0.72 \\
Qwen2.5-VL-72B ~\cite{qwen2024}    & 22.40 & 68.18 & 35.43 & 30.60 & 0.72 & 1.50 & 0.65 \\
Claude-3.7 ~\cite{claude2024}        & 28.40 & 71.06 & 34.77 & 32.50 & 0.85 & 1.16 & 0.70 \\
Qwen-VL-Max ~\cite{qwen2024}       & 30.71 & 68.35 & 35.25 & 33.30 & 0.86 & 1.96 & 0.74 \\
Deepseek V3~\cite{deepseek2025}        & 31.15 & 71.10 & 34.95 & 34.00 & 0.89 & 1.94 & 0.68 \\
GPT-4o     ~\cite{gpt4o2024}        & 34.21 & 73.52 & 35.27 & 30.70 & 0.90 & 1.93 & 1.00 \\
llama-4-maverick ~\cite{meta2025llama4}   & 32.50 & 71.69 & 36.12 & 34.70 & 0.91 & 1.70 & 0.82 \\
claude-sonnet-4~\cite{claude2024}    & 36.73 & 75.76 & 35.30 & 36.60 & \textbf{0.95} & 1.30 & 0.78 \\
\bottomrule
\end{tabular}
}
\end{table*}

\begin{table}[htbp]
\centering
\caption{Performance of findings similarity metrics (BERTScore, ROUGE-L, BLEU-1, METEOR) [in \%]}
\label{tab:scaled-baseline-results2}
\resizebox{\linewidth}{!}{
\begin{tabular}{lccccc}
\toprule
Model & BERTScore (F1) & ROUGE-L & BLEU-1 & METEOR & Diagnosis Accuracy \\
\midrule
Qwen2.5-VL-32B~\cite{qwen2024}    & 70.0 & 12.6 & 7.9  & 20.9 & 7.4  \\
Qwen2.5-VL-7B~\cite{qwen2024}     & 72.7 & 18.6 & 17.4 & 19.4 & 6.5  \\
InternVL3-78B~\cite{internvl3_2025}     & 72.7 & 18.6 & 17.4 & 19.4 & 6.5  \\
Kimi-latest~\cite{kimi2025}       & 73.6 & 21.0 & 16.7 & 21.9 & 8.1  \\
GPT-4O~\cite{gpt4o2024}             & 75.1 & 22.2 & 19.7 & 24.1 & 30.7 \\
Qwen2.5-VL-72B    & 75.0 & 22.9 & 19.7 & 24.1 & 30.6 \\
Claude-3.7 ~\cite{claude2024}        & 74.8 & 22.0 & 18.9 & 23.4 & 32.5 \\
Qwen-VL-Max       & 75.0 & 22.6 & 19.3 & 24.1 & 33.3 \\
Deepseek V3 ~\cite{deepseek2025}       & 74.5 & 22.2 & 20.4 & 22.7 & 34.0 \\
claude-sonnet-4 ~\cite{claude2024} & 75.6 & 22.4 & 17.2 & 26.0 & 36.6 \\
llama-4-maverick ~\cite{meta2025llama4}   & 75.3 & 24.6 & 21.0 & 23.6 & 34.7 \\
\bottomrule
\end{tabular}
}
\end{table}
\section{Experimental Analysis}

We evaluate eleven vision-language models (VLMs), spanning both general-purpose and medical-specialized systems, on the proposed \textbf{MedAtlas benchmark}. The results are summarized in Table~\ref{tab:scaled-baseline-results} (core benchmark) and Table~\ref{tab:scaled-baseline-results2} (findings similarity metrics). This section presents a detailed analysis across four dimensions: overall knowledge reasoning (\emph{Knowledge Acc.}), task-specific performance, cross-metric correlations, and linguistic fidelity in medical report alignment.

\subsection{Overall Knowledge Reasoning Performance}
The \emph{Knowledge Acc.}, derived from Z-score normalization followed by logistic mapping across \emph{Open QA}, \emph{Closed QA}, \emph{Multi-Image Findings}, and \emph{Diagnosis}, offers a unified view of knowledge-grounded multimodal reasoning. As shown in Table~\ref{tab:scaled-baseline-results}, \textbf{Claude-sonnet-4} achieves the highest score (0.95), followed closely by \textbf{LLaMA-4-Maverick} (0.91), \textbf{GPT-4o} (0.90), and \textbf{Deepseek V3} (0.89). These models consistently demonstrate strong general reasoning while retaining sufficient medical grounding.  
In contrast, smaller models such as \textbf{InternVL3-78B} and \textbf{Qwen2.5-VL-7B} record near-zero scores, underscoring the substantial gap between state-of-the-art and early-generation or undertrained systems. Statistical tests (one-way ANOVA, $p < 0.01$) confirm significant separation between these performance tiers.

\subsection{Task-Specific Analysis}
\paragraph{Open vs. Closed QA.}
Although most models achieve reasonable scores in \emph{Closed QA} (e.g., Claude-sonnet-4: 75.76), \emph{Open QA} remains challenging, with a notable drop (Claude-sonnet-4: 36.73). This disparity indicates that while factual recall anchored in fixed knowledge is well addressed, open-ended clinical reasoning involving implicit context remains underexplored.

\paragraph{Multi-Image Findings and Temporal Fusion.}
Performance on \emph{Multi-Image Findings}—computed as the weighted average of BERTScore, ROUGE-L, BLEU-1, and METEOR—strongly correlates with diagnosis accuracy (Pearson $r = 0.82$). Models such as \textbf{LLaMA-4-Maverick} (36.12) excel, suggesting improved temporal fusion and spatial reasoning, whereas \textbf{Qwen2.5-VL-32B} (27.85) lags, indicating insufficient cross-slice integration typical of CT or MRI interpretation workflows.

\paragraph{Diagnosis Accuracy.}
Diagnosis performance mirrors the stratification in \emph{Knowledge Acc.}, with \textbf{Claude-sonnet-4} (36.6), \textbf{Deepseek V3} (34.0), and \textbf{LLaMA-4-Maverick} (34.7) outperforming others. This alignment underscores that diagnostic reasoning benefits from both language fidelity and multimodal fusion, rather than superficial report similarity alone.

\subsection{Findings Similarity and Language Fidelity}
Table~\ref{tab:scaled-baseline-results} (findings similarity) evaluates models on report alignment. \textbf{Claude-sonnet-4} again leads in \emph{BERTScore (75.6)} and \emph{METEOR (26.0)}, reflecting superior semantic alignment with radiology lexicons. Conversely, while \textbf{LLaMA-4-Maverick} achieves the highest \emph{BLEU-1 (21.0)}, lexical overlap alone proves less predictive of diagnostic accuracy, reinforcing the importance of embedding-based semantic metrics over token-level comparisons in clinical text evaluation.

\subsection{Sequential Consistency and Error Propagation Analysis.}
As shown in Table~\ref{tab:scaled-baseline-results}, models with higher \textbf{Knowledge Acc.} generally achieve stronger \textbf{Stage Chain Accuracy (SCA)}, indicating better consistency in sustaining correct reasoning across multi-Round tasks. For example, GPT-4o and Qwen-VL-Max exhibit near-maximal SCA ($\approx$2.0), demonstrating their ability to maintain correctness across consecutive Rounds. In contrast, weaker models (e.g., InternVL3-78B) fail early in the reasoning chain, leading to much lower SCA ($<0.3$).

Similarly, \textbf{Error Propagation Suppression Coefficient (EPSC)} correlates with model robustness: high-performing models such as GPT-4o (EPSC=1.0) show minimal degradation in later Rounds even after early errors, whereas low-performing models (EPSC $<0.3$) exhibit strong error propagation, compounding mistakes across Rounds. These results highlight that advanced VLMs not only improve single-Round accuracy but also exhibit enhanced \emph{sequential reasoning resilience}, a critical property for multi-step medical decision-making.

\subsection{Cross-Metric Correlations and Error Patterns}
Cross-metric analysis highlights several systematic weaknesses.  
First, models with poor \emph{Multi-Image Findings} (e.g., Qwen2.5-VL-7B) display parallel deficits in \emph{Diagnosis}, revealing that limited cross-view integration constrains downstream reasoning.  
Second, models with high \emph{Closed QA} but low \emph{Open QA} (e.g., Qwen2.5-VL-72B) exhibit retrieval-oriented behavior, relying on memorized facts rather than generative inference.  
These patterns suggest that advancing temporal attention and incorporating retrieval-augmented strategies may be critical for further gains.

\subsection{Key Insights}
We draw three primary insights:  
(1) \textbf{Scaling alone is insufficient.} While larger general models like GPT-4o achieve competitive reasoning, specialized pretraining (e.g., Deepseek V3) is crucial for clinical fidelity.  
(2) \textbf{Multi-image reasoning is a bottleneck.} Strong correlation with diagnosis highlights this as an indispensable benchmark axis for future VLM development.  
(3) \textbf{Semantic metrics better capture clinical alignment.} BERTScore and METEOR more closely track diagnostic performance than lexical overlap metrics, suggesting their suitability for medical language evaluation.

\subsection{Conclusion}
Overall, while frontier models approach robust medical reasoning, substantial gaps persist in \emph{open-ended clinical QA, multi-image contextualization, and semantically grounded language generation}. The proposed \textbf{MedAtlas benchmark} thus not only ranks existing systems but also exposes modality-specific limitations, offering a rigorous basis for guiding next-generation \emph{medically grounded vision-language models}.

\bibliography{aaai2026}

\newpage
\newpage
\appendix
\section{Appendix}
A broad array of baseline models was evaluated on \textbf{MedAtlas}. These models include both general vision-language models and those specialized for medical imaging. Unless otherwise noted, publicly available pre-trained weights were used, and models were prompted in a zero-shot fashion. For models requiring specific prompting formats (e.g., an \texttt{<image>} token placeholder or special dialogue format), their official recommended usage was followed.

\subsection{Baseline Models}

In our experiments, we primarily evaluated general multimodal models. Although several multimodal models specialized for medical domains have recently emerged, including LLaVA-Med~\cite{li2023llava}, RADFM~\cite{zhang2023radfm}, Med-PaLM M~\cite{tu2024towards}, we chose not to utilize these domain-specific models due to the following reasons:

Firstly, existing medical multimodal models typically do not support complex interactions involving multiple images, textual information, and multi-turn dialogues. This significantly limits their applicability to real-world scenarios requiring intricate multimodal interactions. For instance, during our evaluation of LLaVA-Med, we found that the model only supports single-image and text interactions, lacking the capability for effective cross-modal information fusion across multiple images. Secondly, these medical-specific models often possess limited context windows, making them inadequate for handling longer sequences of multimodal inputs. Additionally, we observed that LLaVA-Med was incapable of reliably adhering to instructions during the tests, demonstrating significant performance limitations. Furthermore, Med-PaLM M is currently still under research development and lacks publicly available API interfaces, which impedes its deployment in practical applications and large-scale evaluations.

Considering these constraints, we opted to employ general multimodal models as our primary testing tools. A comprehensive suite of vision-language models (VLMs) was evaluated on \textbf{MedAtlas}, covering both general-purpose and domain-specialized architectures. All models represent the latest available versions released in 2025, ensuring a fair comparison with state-of-the-art methods. Unless otherwise specified, publicly available pre-trained weights were employed, and zero-shot prompting was performed, thus ensuring the effectiveness, generality, and adaptability of our experiments in complex multimodal interaction scenarios.

These models were trained on large-scale general-domain corpora without medical-specific fine-tuning:
\begin{itemize}
    \item \textbf{Claude-Sonnet-4}: Anthropic's most advanced Claude model, offering enhanced vision capabilities and strong reasoning performance.
    \item \textbf{GPT-4o}: OpenAI’s flagship multi-modal model integrating high-resolution image understanding and robust language reasoning.  
    \item \textbf{Qwen-VL-Max}: Alibaba’s Qwen series model with superior visual comprehension and extended context handling.
    \item \textbf{Deepseek V3}: A high-performing multi-modal model excelling in both Chinese and English tasks.
    \item \textbf{Claude-3.7}: A vision-enabled model from Anthropic’s Claude 3 series, noted for its alignment and interpretability.
    \item \textbf{Kimi-latest}: Moonshot AI’s conversational assistant optimized for image-grounded reasoning.
    \item \textbf{InternVL3-78B}: The latest release in the InternVL series, built for large-scale visual-linguistic understanding.
    \item \textbf{Qwen2.5-VL-72B/32B/7B}: The newest generation of Qwen2.5-VL models, supporting multi-image inputs and improved visual grounding.
\end{itemize}

Each model was prompted in a Round-wise manner. For dialogue models, prompts followed the structure: “History: [text]. Image: [image]. Question 1: [text]? Answer 1: … Question 2: [text]? Answer 2: …”. Non-dialogue models were prompted with one question at a time, incorporating relevant context. For models such as GPT-4o that accept vision input, images were provided sequentially, and conversation history was carried forward within the prompt.

\textbf{Evaluation Protocol}: All models were evaluated on the \textbf{MedAtlas} test set under standardized conditions. Generation lengths were constrained for consistency. Greedy decoding was used for closed questions to prevent extraneous text generation, while a low temperature was applied to open questions to enable descriptive yet reliable answers. Ground truth answers and test data were not exposed during inference. For models requiring special instruction prompts (e.g., “You are a helpful medical assistant…”), the required preamble was included. 

Each model was assessed on the tasks defined in the main paper, yielding a range of metrics: overall accuracy, BLEU, Stage Chain Accuracy, EPSC, and more. Results are compiled in Table~2, which reports open QA accuracy, closed QA accuracy, multi-image question accuracy, findings generation F1, knowledge aspect accuracy, overall Stage Chain Accuracy, and EPSC. The best-performing model for each metric is boldfaced, and the second-best is underlined.

Additionally, Figure~2 illustrates Stage Chain Accuracy versus overall accuracy, highlighting that models with similar accuracy may differ in consistency.

All experiments were conducted on an internal cluster equipped with A800 GPUs. The largest models (e.g., GPT-4o) were accessed via cloud APIs. Context lengths were sufficient for all models, though a small fraction ($<5\%$) of very long dialogues had to be truncated for a model limited to 1024 tokens. Mode

\subsection{Data Sourcing and Curation}

\textbf{Data Source and Construction:} The \textbf{MedAtlas} dataset is derived from publicly available medical case repositories and expert clinical case collections. Cases were collected from online radiology education platforms and forums where experienced radiologists share real-world patient scenarios, including imaging studies and discussions of findings. 

The curation process followed several steps. Cases were first gathered to ensure coverage of a broad range of conditions (e.g., musculoskeletal injuries, neurological disorders, cardiovascular issues) and imaging modalities (X-ray, CT, MRI, ultrasound, PET, angiography). Each case was then restructured into the \textbf{MedAtlas} format by segmenting it into distinct “Rounds” reflecting the natural clinical progression described in the source material. For example, a case described as: initial presentation and X-ray findings, followed by MRI findings, and concluding with a diagnosis, was mapped to Round I, Round II, and so forth. Clinical history (patient age, sex, symptoms, prior history) was extracted as a prelude to Round I, and all relevant medical images were included for each Round. When multiple images per Round were available (e.g., anteroposterior and lateral X-rays, or multiple MRI slices), all images pertinent to the questions in that Round were retained.

Question-answer (QA) pairs were then formulated for each Round. When source material contained explicit teaching questions (e.g., “What do you see on this image?”), these were retained. Otherwise, clinically relevant questions typical of medical training or diagnostic reasoning were generated, and corresponding answers were derived from the accompanying case explanations or reports. All QA pairs underwent review by licensed medical professionals to ensure clinical correctness and logical coherence across Rounds. The questions encompass multiple types: (i) identification (e.g., “What abnormality is seen on the X-ray?”), (ii) reasoning (e.g., “Given the X-ray findings, what is the next best imaging step?”), (iii) differential diagnosis (e.g., “What are two possible diagnoses at this point?”), and (iv) decision-making (e.g., “What is the most likely diagnosis?” or “What treatment is indicated?”). Later questions are intentionally dependent on earlier Rounds. For instance, a Round II question such as “The MRI shows a tear of which ligament?” presupposes recognition in Round I that MRI is indicated and necessitates interpretation of subsequent imaging. 

Each case includes ground-truth answers to all questions, as well as a \emph{Findings} section summarizing image observations for each Round, and a \emph{Diagnosis \& Discussion} section providing the confirmed diagnosis and a structured explanation incorporating five medical knowledge aspects: pathophysiology, epidemiology, clinical features, imaging features, and treatment. These annotations serve both as references for evaluation and as potential extended context for models capable of processing long-form textual inputs.

Figure~\ref{fig:example} illustrates the dataset’s structure, presenting an example QA chain from \textbf{MedAtlas}. The figure shows the clinical history, followed by Round I images and questions, then Round II with follow-up imaging and questions, culminating in the diagnosis. This example highlights the interleaving of textual and visual information, requiring models to interpret images and maintain contextual memory to address later questions.

\subsection{Core Characteristics}

\textbf{Multi-round \& Stage Dependency:} Subsequent questions frequently depend on prior answers, requiring models to utilize conversation history and maintain contextual continuity, including handling references such as pronouns (e.g., “this finding”).

\textbf{Multi-modality:} \textbf{MedAtlas} spans multiple imaging modalities (X-ray, CT, MRI, ultrasound, PET, mammography, angiography), often within a single case. This diversity necessitates reasoning across heterogeneous visual data and exceeds the modality range of prior benchmarks such as VQA-RAD~\cite{lau2022vqa_rad} and SLAKE~\cite{liu2021slake}.

\textbf{Multi-image per Round:} Certain questions involve simultaneous reasoning over multiple images within a Round (e.g., comparing two X-ray views or synthesizing information across MRI slices), akin to NLVR2-style multi-image reasoning~\cite{suhr2019nlvr2}, but adapted to medical imaging complexity.

\textbf{Clinical Reasoning Chain:} The QA structure mirrors clinical reasoning stages, progressing from early observation to intermediate synthesis and finally to diagnostic or treatment decisions. This design evaluates multi-hop reasoning and the integration of medical knowledge throughout the case.

\textbf{Disease Coverage:} \textbf{MedAtlas} encompasses over 50 diagnoses spanning multiple specialties (musculoskeletal, neurological, thoracic, abdominal, pediatric, etc.), balancing common conditions (e.g., appendicitis, pneumonia) with less frequent but clinically significant entities. Although educational cases often emphasize unusual findings, representative routine cases are included to maintain clinical realism and diversity.
\subsection{Image and text consumption}

\begin{figure*}[t]
    \centering
    \begin{subfigure}{0.45\linewidth}
        \includegraphics[width=\linewidth]{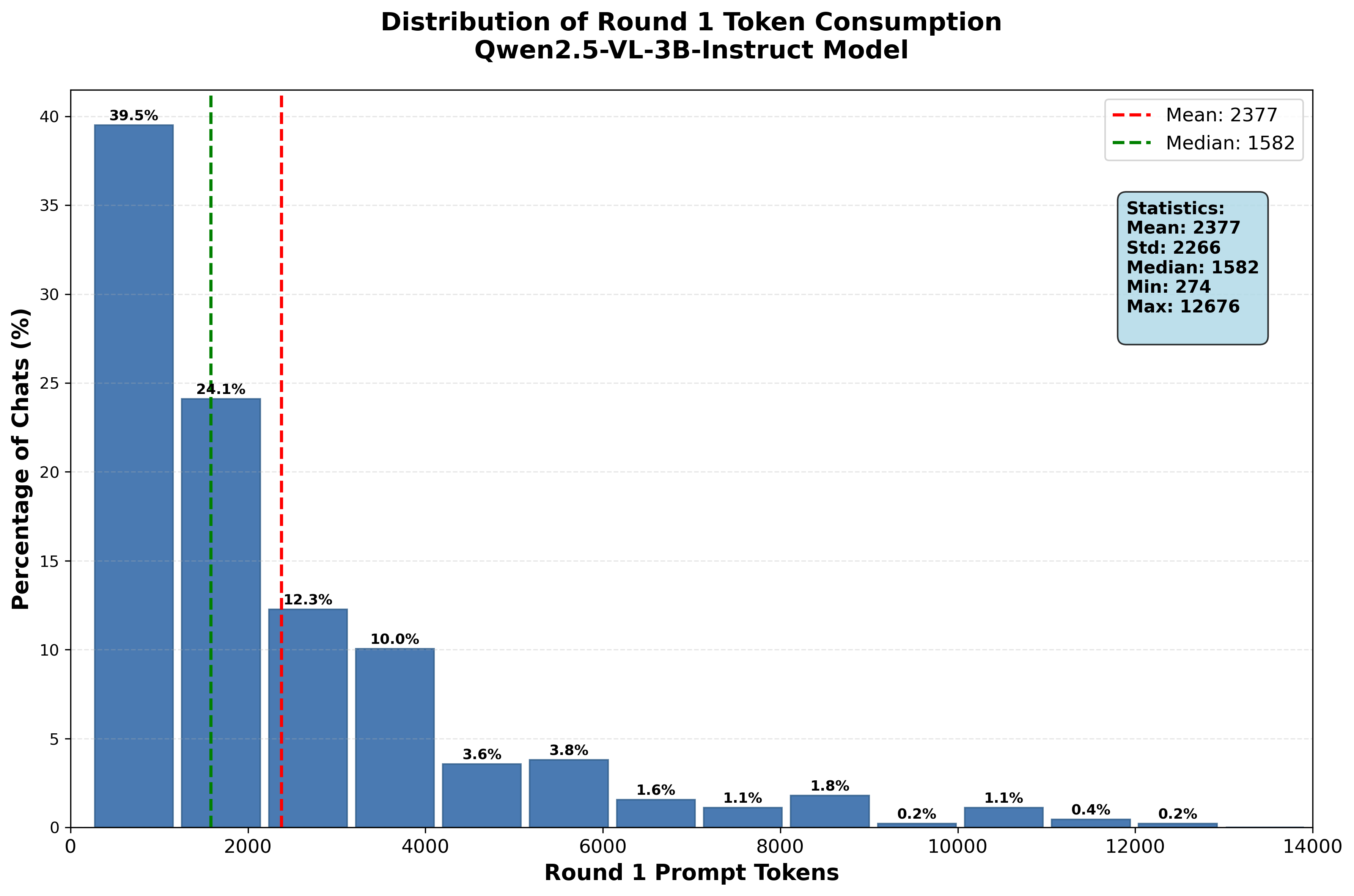}
        \caption{Distribution of token consumption in Round~1 prompts.}
    \end{subfigure}
    \hspace{0.05\linewidth}
    \begin{subfigure}{0.45\linewidth}
        \includegraphics[width=\linewidth]{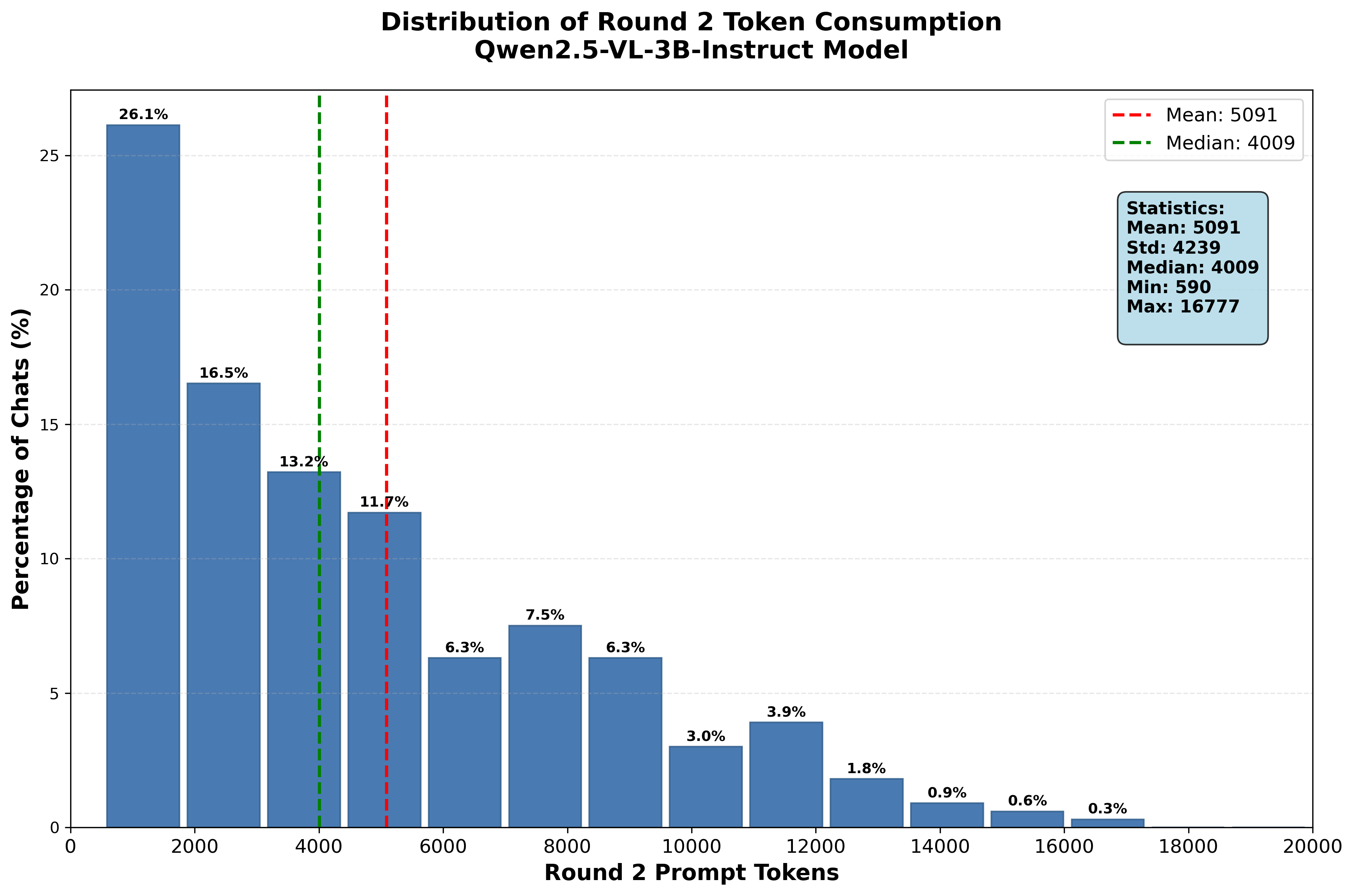}
        \caption{Distribution of token consumption in Round~2 prompts}
    \end{subfigure}

    \vspace{0.5em}

    \begin{subfigure}{0.45\linewidth}
        \includegraphics[width=\linewidth]{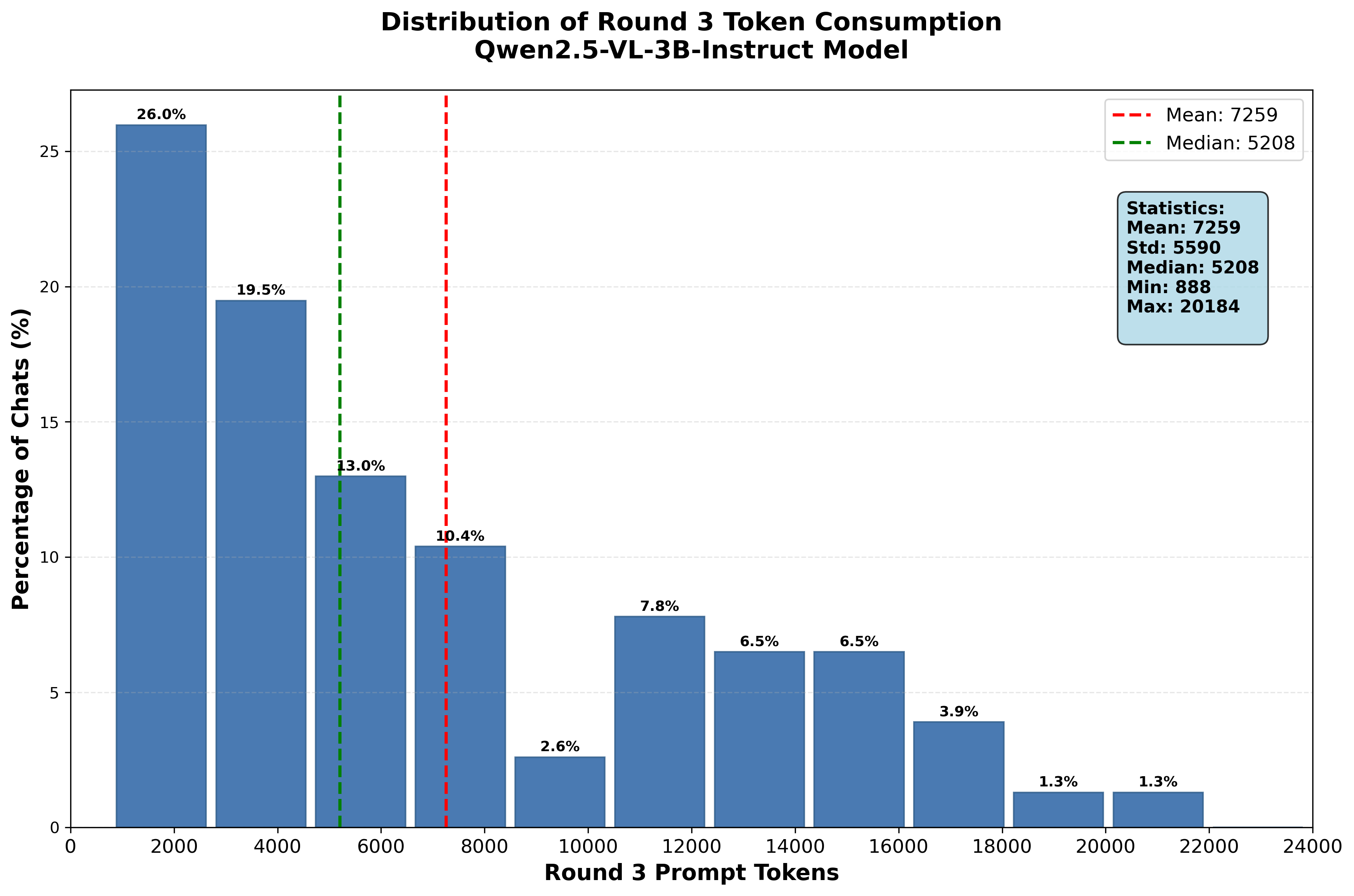}
        \caption{Distribution of token consumption in Round~3 prompts.}
    \end{subfigure}
    \hspace{0.05\linewidth}
    \begin{subfigure}{0.45\linewidth}
        \includegraphics[width=\linewidth]{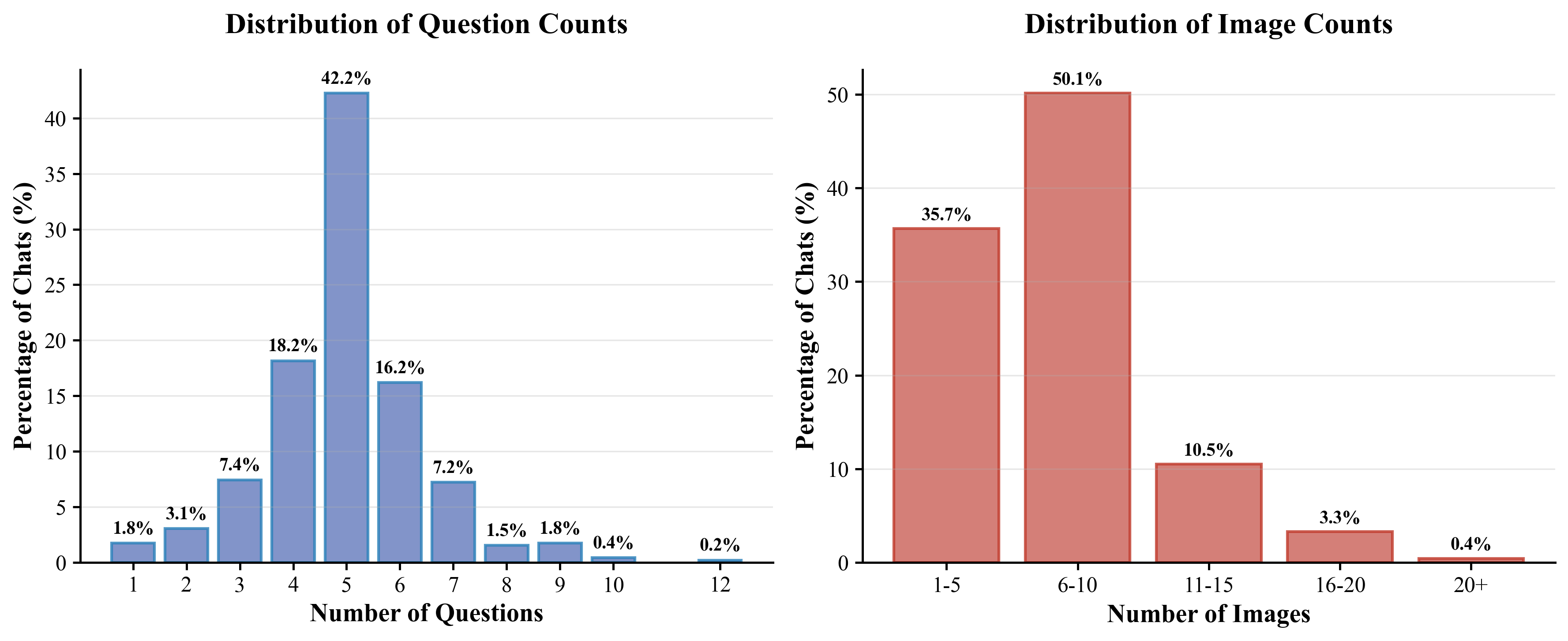}
        \caption{Distribution of question and image counts in the Multiple-round tasks.}
    \end{subfigure}
    
 \caption{Statistics of prompt token usage and dataset composition. 
    (a--c) Token consumption distributions across Rounds~1, 2, and 3, respectively, highlighting mean and median values for each round.
    (d) Dataset statistics showing the distribution of question counts per case and image counts per case.}
    \label{fig:token_and_data_stats}
\end{figure*}

As shown in Figure~\ref{fig:token_and_data_stats}, the token consumption for model prompts exhibits a clear progression across interaction rounds. 
Specifically, the median token usage increases from Round~1 (1,582 tokens) to Round~2 (4,099 tokens) and Round~3 (5,208 tokens), reflecting the growing contextual length required for multi-round reasoning. 
The distributions demonstrate that a majority of cases concentrate near the lower token ranges, with long-tail instances corresponding to complex multi-phase cases.

Additionally, the dataset composition statistics in Figure~\ref{fig:token_and_data_stats}(d) show that most clinical cases contain 4--7 questions and 1--6 images. 
This aligns with typical diagnostic workflows in radiology education, where a single case often involves multiple images across different views or modalities and sequential question-answer reasoning. 
These distributions highlight both the efficiency challenges posed by increasing token demands and the realistic complexity of the \textbf{MedAtlas} benchmark, underscoring the necessity for models to manage multi-turn dialogue while maintaining image-grounded reasoning.

\begin{table}[ht]
    \small
    \caption{The prompt for multi-round and multi-modal closed-ended responses is as follows:}
    \centering
    \begin{tcolorbox}[colback=gray!10, colframe=black, width=0.48\textwidth, arc=3mm, boxrule=0.5mm]
        You are a doctor. Please answer each question in sequence based on the provided medical images and text.\\[1mm]

        \textbf{Use the format:}\\
        Question\textless number\textgreater\\
        Correct: \textless A--Z\textgreater\\
        Explanation: \textless your explanation here\textgreater
    \end{tcolorbox}
    \label{tab:prompt_doctor}
\end{table}

\begin{table}[ht]
    \small
    \caption{The prompt design for open-ended question answering (Open QA). Only the English version is shown here; prompts in other languages follow a similar structure.}
    \centering
    \begin{tcolorbox}[colback=gray!10, colframe=black, width=0.48\textwidth, arc=3mm, boxrule=0.5mm]
        You are a doctor. Please answer each question in sequence based on the provided medical images and text \textbf{without explanation}.\\[1mm]

        \textbf{Use the format:}\\
        Question\textless number\textgreater\\
        Brief Answer: \textless answer\textgreater
    \end{tcolorbox}
    \label{tab:prompt_openqa}

\end{table}

% \begin{table*}[ht]
%     \small
%     \caption{The prompt design for multi-image reasoning and comprehensive disease diagnosis. Only the English version is shown here; prompts in other languages follow the same structure.}
%     \centering
%     \begin{tcolorbox}[colback=gray!10, colframe=black, width=0.48\textwidth, arc=3mm, boxrule=0.5mm]
%         You are a medical expert. Please describe the findings in detail based on the uploaded medical images and choose the most likely option.\\[1mm]

%         \textbf{Use the format:}\\
%         Findings: \textless description\textgreater\\
%         Option: \textless option\textgreater
%     \end{tcolorbox}
%     \label{tab:prompt_multimage}
% \end{table*}

\clearpage
\clearpage
\begin{figure*}[ht]
\begin{tcolorbox}[
  enhanced jigsaw,
  colback=white, colframe=black!60,
  title=Case Study – Vascular Imaging in Migraine Patient,
  width=\textwidth,
  enlarge top initially by=0mm,
  enlarge bottom finally by=0mm,
  fonttitle=\bfseries,
  fontupper=\normalsize
]

\textbf{Clinical History:} \textit{A 46-year-old man presented with a history of migraines with photophobia, phonophobia, and nausea. The photophobia often served as a trigger for his headache episodes; each episode typically lasted two to three hours and occurred four times per month. There were no neurological deficits found on exam.}

\vspace{0.8em}
\begin{tcolorbox}[colback=gray!5, colframe=black!60, title=Phase I – Provided Imaging, breakable]
\begin{center}
\includegraphics[height=2.2cm]{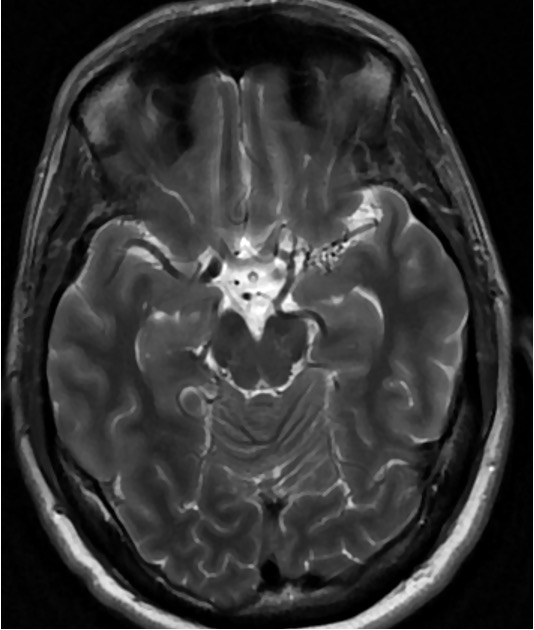}
\hspace{1em}
\includegraphics[height=2.2cm]{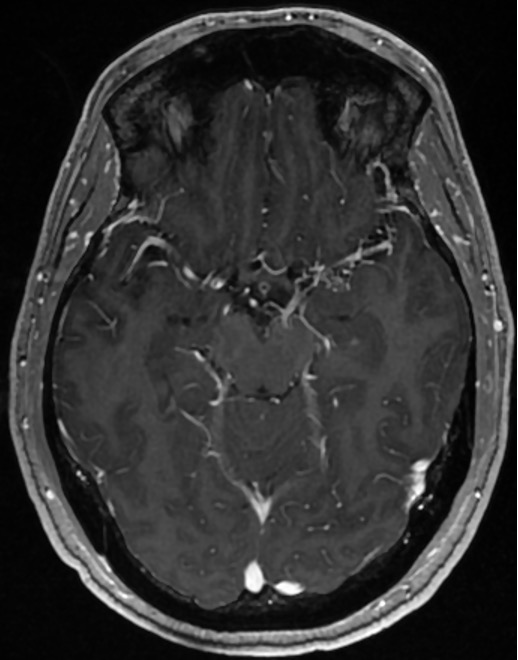}
\hspace{1em}
\includegraphics[height=2.2cm]{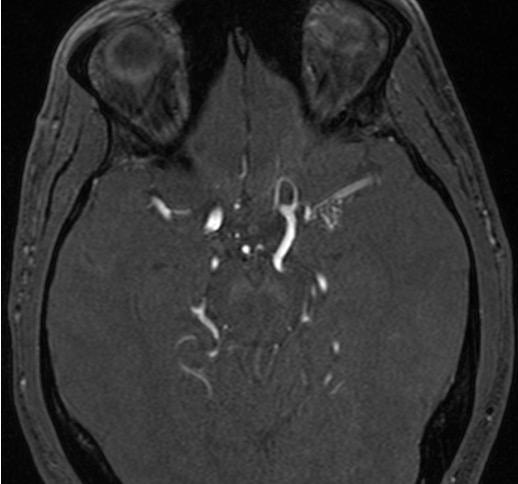}
\hspace{1em}
\includegraphics[height=2.2cm]{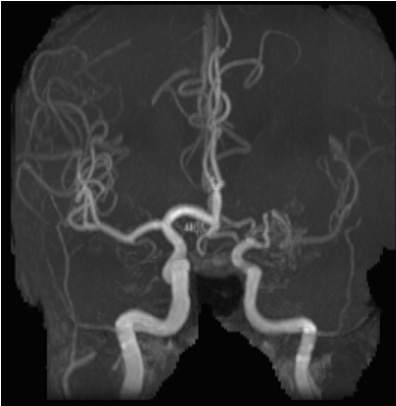}
\end{center}

\vspace{1em}
\textbf{Q1:} There is severe stenosis of the proximal left M1 segment. \\
\textbf{Options:} \\
A. TRUE \\
B. FALSE \\[0.5em]
\textbf{Prediction: A. TRUE \hfill \textcolor{green!50!black}{\textbf{  Correct}}} \\
\textbf{Ground Truth: A} \\
\textbf{Explanation:} The images suggest severe stenosis of the proximal left M1 segment, supported by the abnormal appearance and flow in the left middle cerebral artery region.

\vspace{1em}
\textbf{Q2:} Which area reveals an abnormality? \\
\textbf{Options:} \\
A. Right Sylvian fissure \\
B. Substantia nigra \\
C. Left Sylvian fissure \\
D. Genu of corpus collosum \\[0.5em]
\textbf{Prediction: C. Left Sylvian fissure \hfill \textcolor{green!50!black}{\textbf{  Correct}}} \\
\textbf{Ground Truth: C} \\
\textbf{Explanation:} The abnormality appears in the left Sylvian fissure, consistent with the described stenosis in the left middle cerebral artery territory.

\vspace{1em}
\textbf{Q3:} What is shown on patient imaging? \\
\textbf{Options:} \\
A. Collateral vessels \\
B. Asymmetric flow voids \\
C. Middle cerebral artery (MCA) stenosis \\
D. All of the above \\[0.5em]
\textbf{Prediction: D. All of the above \hfill \textcolor{green!50!black}{\textbf{  Correct}}} \\
\textbf{Ground Truth: D} \\
\textbf{Explanation:} The images show signs of collateral vessels, asymmetric flow voids, and MCA stenosis, indicating adaptive changes due to vascular compromise.

\end{tcolorbox}

\end{tcolorbox}
\vspace{0.5em}
\caption{ VQA Case Study – Vascular Imaging in Migraine Patient}
\label{fig:case-vascular-migraine}
\end{figure*}

\clearpage

\begin{figure*}[ht]
\begin{tcolorbox}[
  enhanced jigsaw,
  colback=white, colframe=black!60,
  title=Case Study – Pediatric Necrotizing Pancreatitis,
  width=\textwidth,
  enlarge top initially by=0mm,
  enlarge bottom finally by=0mm,
  fonttitle=\bfseries,
  fontupper=\normalsize
]

\textbf{Clinical History:} \textit{A 10-year-old boy presented with complaints of abdominal pain in the epigastric region radiating to the back associated with multiple episodes of vomiting in the last three days. An initial ultrasonography of the abdomen done elsewhere at a clinic was normal at presentation.
In view of progressively increasing upper abdominal pain and vomiting, the child was brought to our institute, a tertiary care center, for further workup. A repeat ultrasonography was performed on day 4 and revealed a mildly bulky pancreas and fluid in the lesser sac and pelvis. The complete blood picture and liver function tests were normal. The serum amylase (596 IU/L) and serum lipase (396 IU/L) values were elevated. A contrast-enhanced CT (CECT) scan of the abdomen was performed.}

\vspace{0.8em}
\begin{tcolorbox}[colback=gray!5, colframe=black!60, title=Phase I – Provided Imaging, breakable]
\begin{center}
\includegraphics[height=2.2cm]{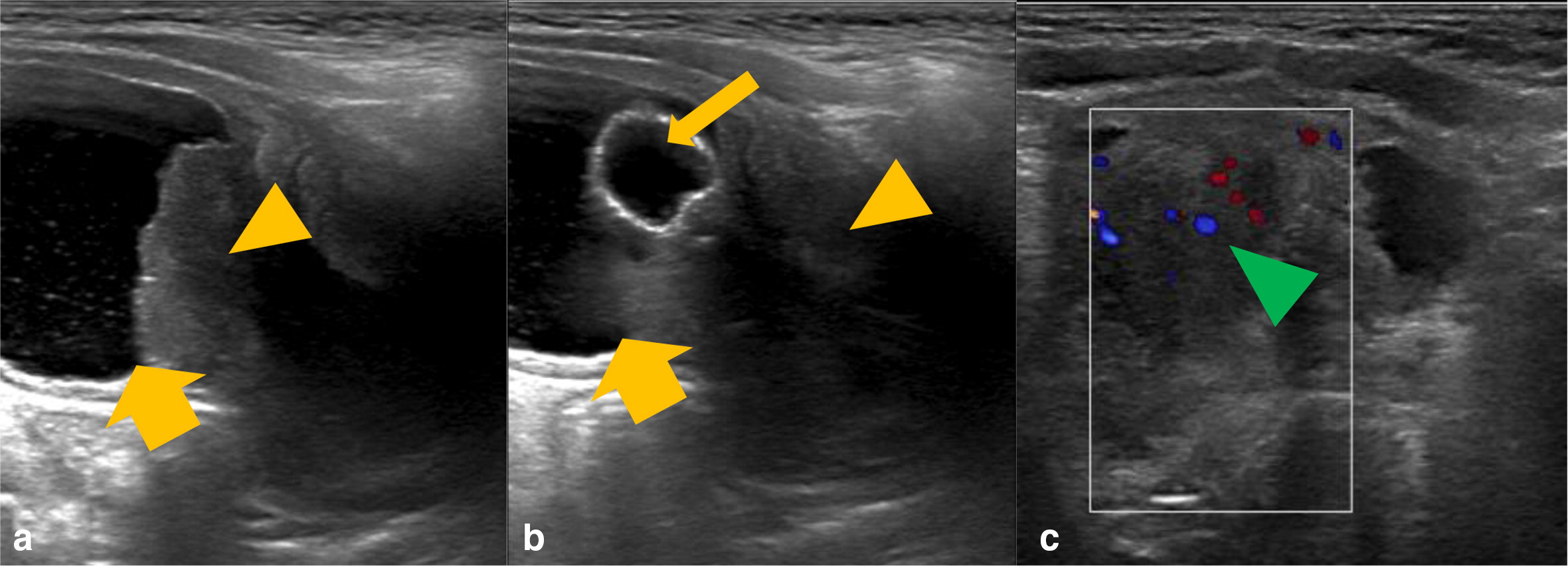}
\hspace{1em}
\includegraphics[height=2.2cm]{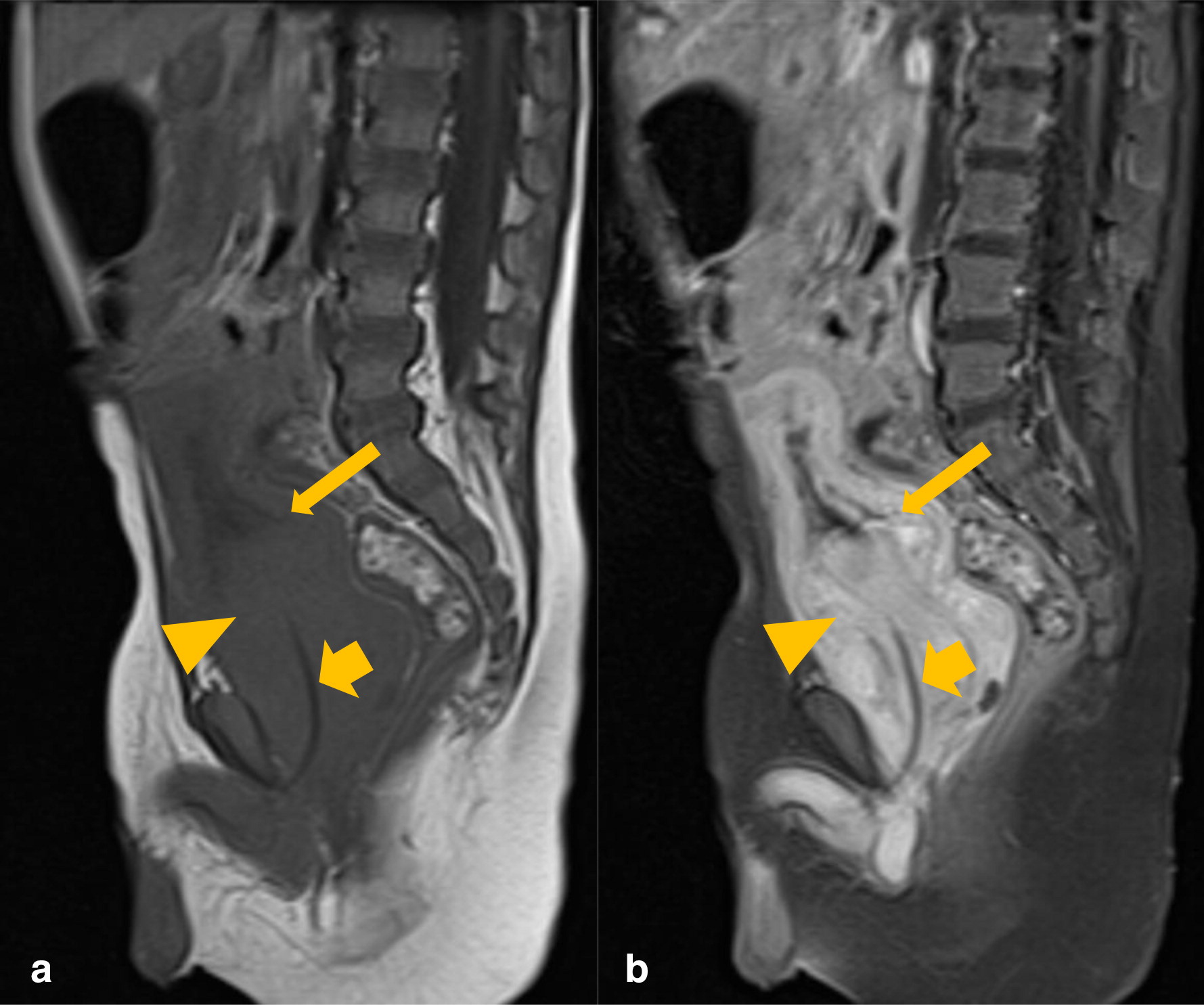}
\hspace{1em}
\includegraphics[height=2.2cm]{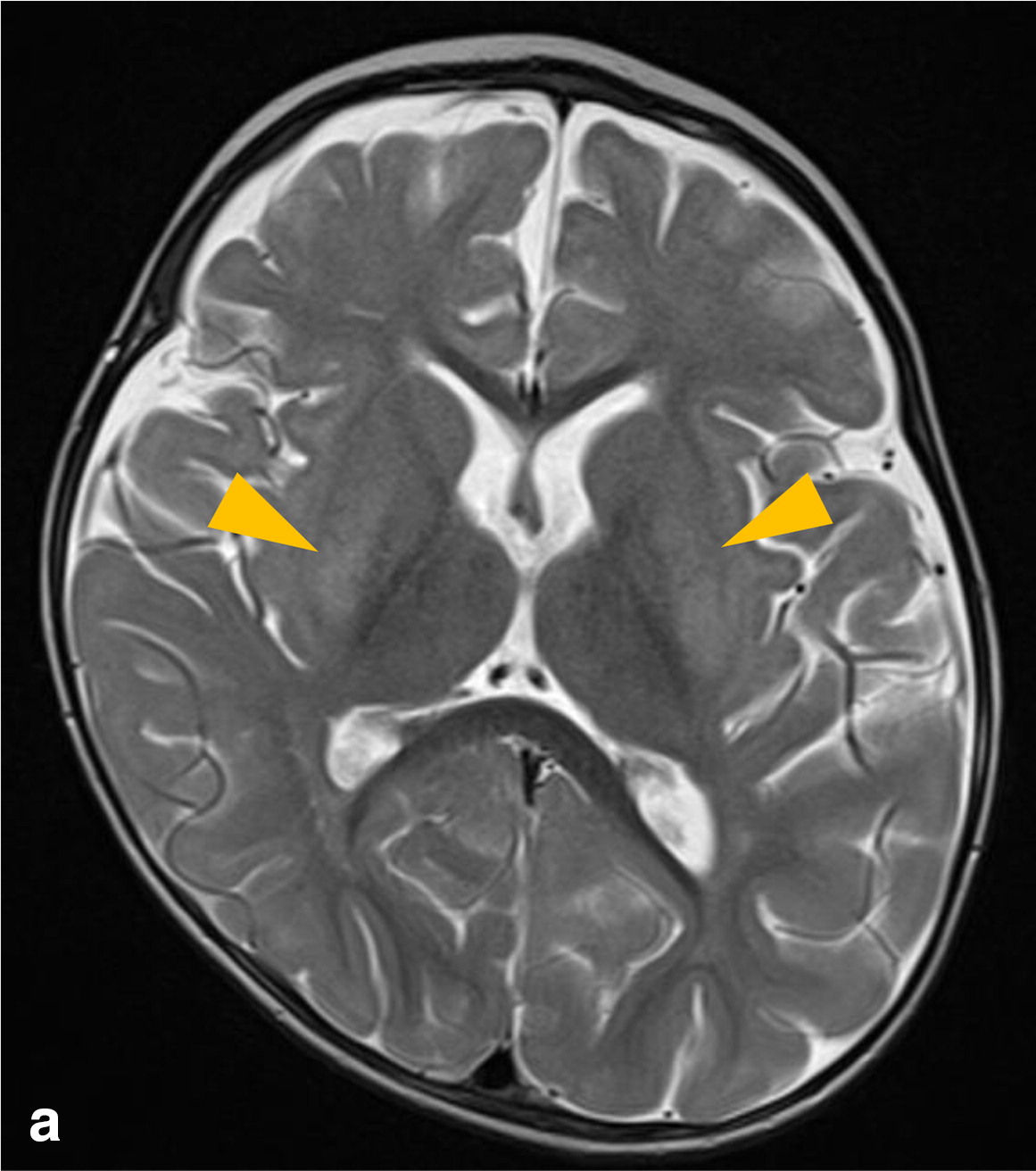}
\end{center}

\vspace{1em}
\textbf{Q1:} What best describes the findings? \\
\textbf{Options:} \\
A. Cystic neoplasm of the pancreas \\
B. Celiac artery thrombus \\
C. Acute cholecystitis \\
D. Necrotizing pancreatitis \\[0.5em]
\textbf{Prediction: D. Necrotizing pancreatitis \hfill \textcolor{green!50!black}{\textbf{  Correct}}} \\
\textbf{Ground Truth: D} \\
\textbf{Explanation:} The findings from the CT images show evidence of a bulky pancreas with fluid collections around it, consistent with necrotizing pancreatitis. The elevated serum amylase and lipase levels support this diagnosis.

\vspace{1em}
\textbf{Q2:} What is the likely cause of the patient’s condition as inferred from the above images? \\
\textbf{Options:} \\
A. Alcoholism \\
B. A penetrating foreign body \\
C. Gallstone \\
D. Congenital anomaly \\[0.5em]
\textbf{Prediction: D. Congenital anomaly \hfill \textcolor{red}{\textbf{  Incorrect}}} \\
\textbf{Ground Truth: B} \\
\textbf{Explanation:} The likely cause of this 10-year-old boy’s acute pancreatitis, given his age and without evidence of alcohol use or gallstones, is a congenital anomaly, such as pancreatic divisum. This can lead to recurrent episodes of pancreatitis in children.

\end{tcolorbox}

\end{tcolorbox}
\vspace{0.5em}
\caption{VQA Case Study – Pediatric Necrotizing Pancreatitis}
\label{fig:case-peds-pancreatitis}
\end{figure*}

\clearpage

\begin{figure*}[ht]
\begin{tcolorbox}[
  enhanced jigsaw,
  colback=white, colframe=black!60,
  title=Case Study – Acute Tonsillar Herniation with Cytotoxic Edema,
  width=\textwidth,
  enlarge top initially by=0mm,
  enlarge bottom finally by=0mm,
  fonttitle=\bfseries,
  fontupper=\normalsize
]

\textbf{Clinical History:} \textit{A 10-year-old girl presented to the emergency room in a semiunconscious condition. She had been gradually losing consciousness and awareness since the previous night. Her parents also mentioned occasional complaints of headache and dizziness. A CT was immediately advised on suspicion of an intracranial pathology as the cause of the sudden loss of consciousness.}

\vspace{0.8em}
\begin{tcolorbox}[colback=gray!5, colframe=black!60, title=Phase I – Provided Imaging, breakable]
\begin{center}
\includegraphics[height=2.2cm]{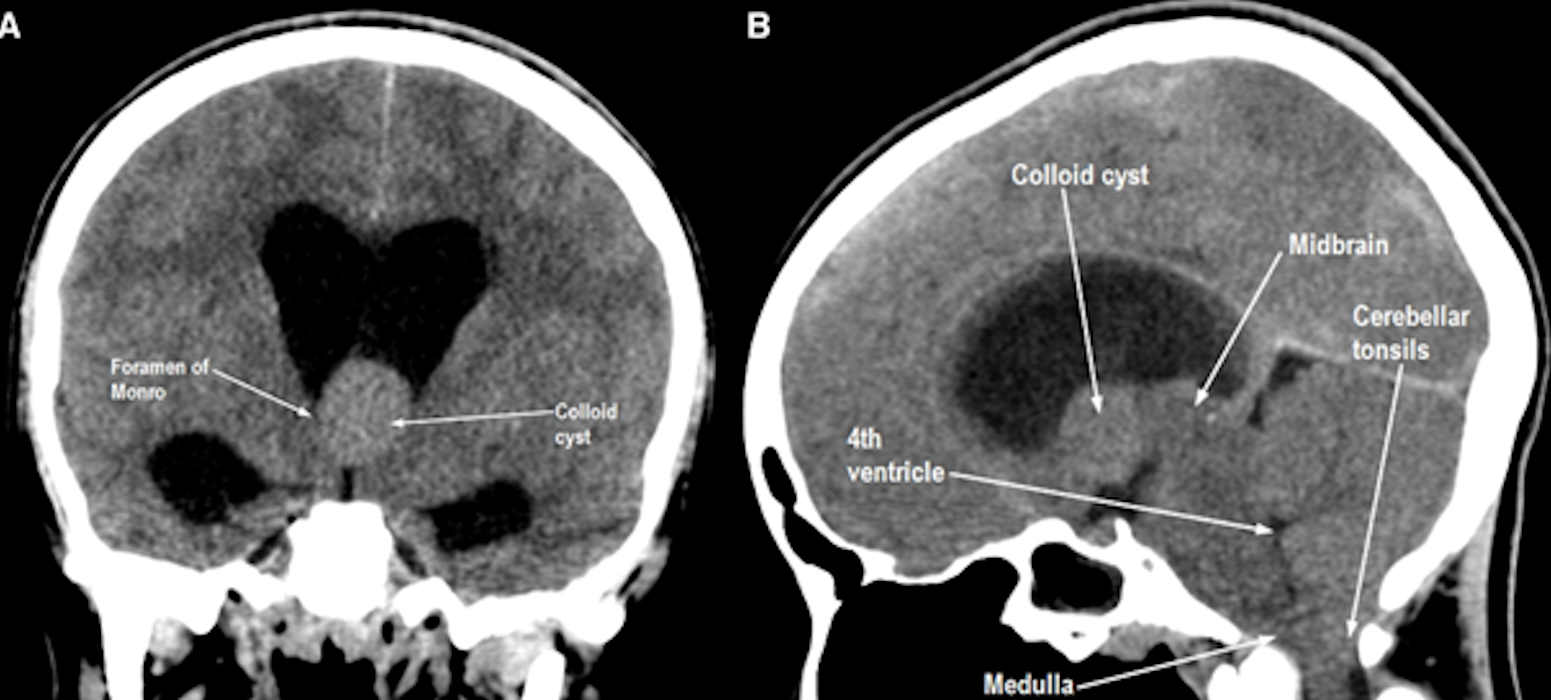}
\end{center}

\vspace{1em}
\textbf{Q1:} What does the accentuated gray-white contrast (as seen on the coronal image) signify in the context of this CT image? \\
\textbf{Options:} \\
A. Vasogenic edema \\
B. Cytotoxic edema \\
C. White matter dystrophy \\
D. Ischemic change \\[0.5em]
\textbf{Prediction: B. Cytotoxic edema \hfill \textcolor{red}{\textbf{  Incorrect}}} \\
\textbf{Ground Truth: A} \\
\textbf{Explanation:} Cytotoxic edema causes a loss of the normal gray-white differentiation due to cellular injury and swelling, as seen in this CT image.

\vspace{1em}
\textbf{Q2:} Up to how far below the level of foramen magnum can the cerebellar tonsils be located normally, i.e., without herniation? \\
\textbf{Options:} \\
A. 3 mm \\
B. 4 mm \\
C. 6 mm \\
D. 1 mm \\[0.5em]
\textbf{Prediction: A. 3 mm \hfill \textcolor{green!50!black}{\textbf{  Correct}}} \\
\textbf{Ground Truth: A} \\
\textbf{Explanation:} Normally, the cerebellar tonsils can be located up to 3 mm below the foramen magnum without being considered herniation.

\vspace{1em}
\textbf{Q3:} Which of the findings on these images had the most lethal effect and ultimately caused the death of the patient? \\
\textbf{Options:} \\
A. Tonsillar herniation \\
B. Brainstem herniation \\
C. Brainstem edema \\
D. Cerebral edema \\[0.5em]
\textbf{Prediction: A. Tonsillar herniation \hfill \textcolor{red}{\textbf{  Incorrect}}} \\
\textbf{Ground Truth: C} \\
\textbf{Explanation:} Tonsillar herniation is life-threatening as it can compress the brainstem and vital structures, causing rapid deterioration and potentially resulting in death.

\end{tcolorbox}

\end{tcolorbox}
\vspace{0.5em}
\caption{ VQA Case Study – Acute Tonsillar Herniation with Cytotoxic Edema}
\label{fig:case-tonsillar-herniation}
\end{figure*}

\clearpage

\begin{figure*}[ht]
\begin{tcolorbox}[
  enhanced jigsaw,
  colback=white, colframe=black!60,
  title=Case Study – Infant Subdural Hemorrhage Following Fall,
  width=\textwidth,
  enlarge top initially by=0mm,
  enlarge bottom finally by=0mm,
  fonttitle=\bfseries,
  fontupper=\normalsize
]

\textbf{Clinical History:} \textit{A 12-week-old infant boy presented to the emergency department following a fall. The boy's parents reported that the infant experienced three episodes of vomiting and was irritable. CT images are shown below.}

\vspace{0.8em}
\begin{tcolorbox}[colback=gray!5, colframe=black!60, title=Phase I – Provided Imaging, breakable]
\begin{center}
\includegraphics[height=2.2cm]{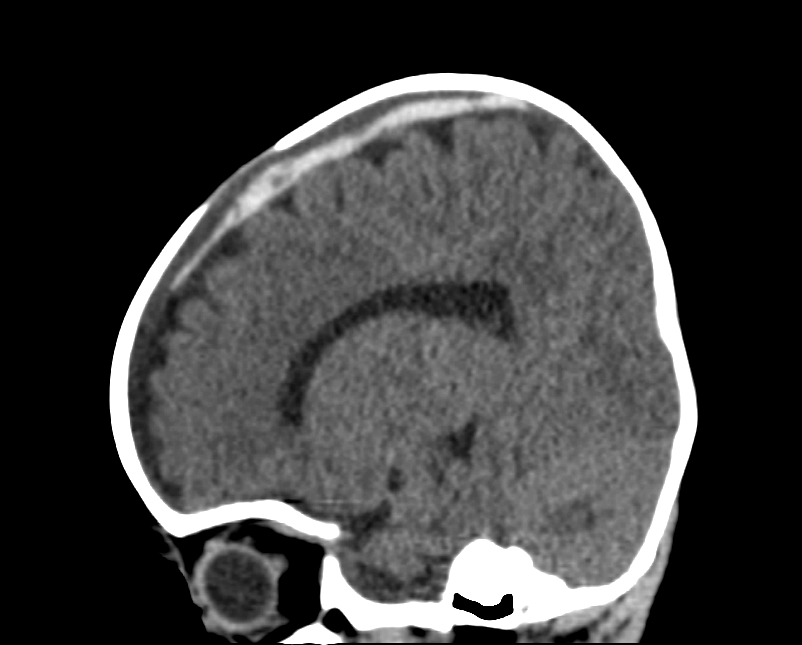}
\hspace{1em}
\includegraphics[height=2.2cm]{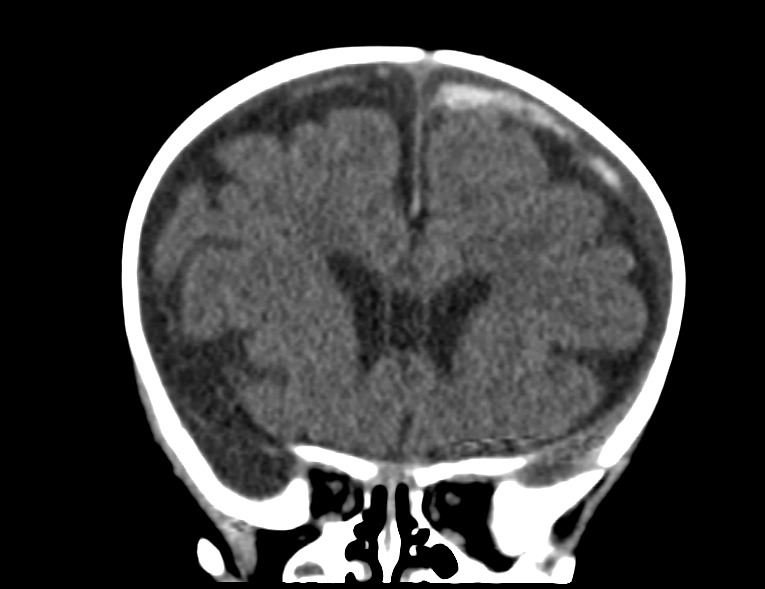}
\hspace{1em}
\includegraphics[height=2.2cm]{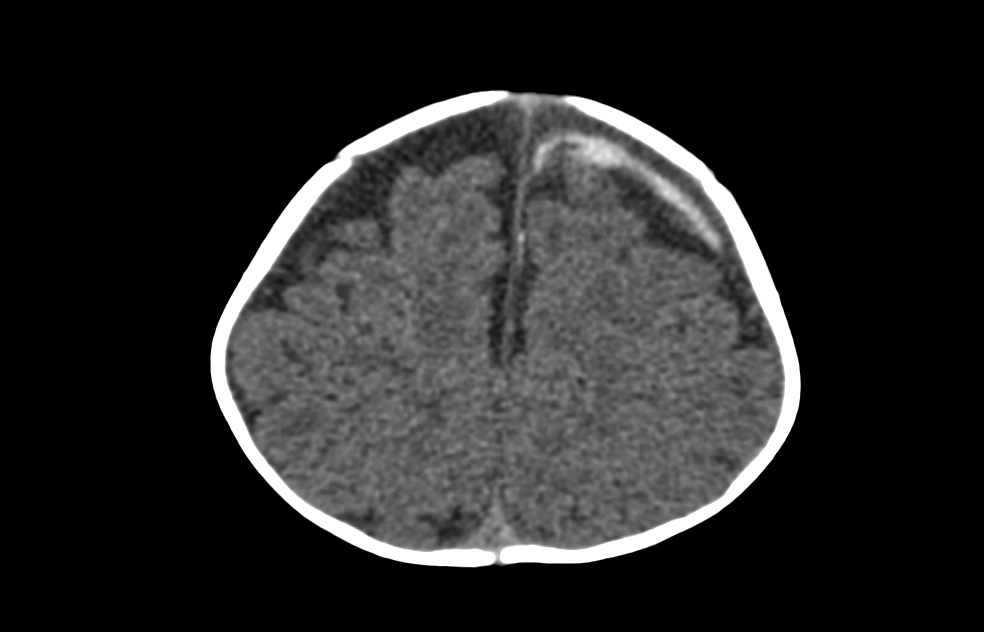}
\end{center}

\vspace{1em}
\textbf{Q1:} What imaging modality should be used next? \\
\textbf{Options:} \\
A. MRI \\
B. X-ray \\
C. PET scan \\
D. Ultrasound \\[0.5em]
\textbf{Prediction: A. MRI \hfill \textcolor{green!50!black}{\textbf{  Correct}}} \\
\textbf{Ground Truth: A} \\
\textbf{Explanation:} The CT images of the infant's brain appear to show signs of a subdural hemorrhage. An MRI is the next best imaging modality to provide a more detailed view of the brain structures and to assess for any additional injuries or underlying conditions, as it offers more detailed brain imaging.

\vspace{1em}
\textbf{Q2:} Which specialty should be consulted? \\
\textbf{Options:} \\
A. Neurosurgery and neurology \\
B. Orthopedics \\
C. Dermatology \\
D. Otolaryngology \\[0.5em]
\textbf{Prediction: A. Neurosurgery and neurology \hfill \textcolor{green!50!black}{\textbf{  Correct}}} \\
\textbf{Ground Truth: A} \\
\textbf{Explanation:} Given the indication of a possible subdural hemorrhage in these CT images, consultation with neurosurgery and neurology would be appropriate. These specialties are equipped to handle conditions involving the brain such as hemorrhages and will be able to provide the necessary management and care.

\vspace{1em}
\textbf{Q3:} What is visible in the CT images? \\
\textbf{Options:} \\
A. Subdural hemorrhage \\
B. Stroke \\
C. Brain mass \\
D. Aneurysm \\[0.5em]
\textbf{Prediction: A. Subdural hemorrhage \hfill \textcolor{green!50!black}{\textbf{  Correct}}} \\
\textbf{Ground Truth: A} \\
\textbf{Explanation:} The CT images suggest the presence of a subdural hemorrhage, which typically appears as a crescent-shaped collection of blood between the dura and the brain surface. The pattern and location of increased density in the CT images support this diagnosis, which is common in infants with a history of traumatic injury.

\end{tcolorbox}

\end{tcolorbox}
\vspace{0.5em}
\caption{ VQA Case Study – Infant Subdural Hemorrhage Following Fall}
\label{fig:case-subdural-hemorrhage-infant}
\end{figure*}

\clearpage

\begin{figure*}[ht]
\begin{tcolorbox}[
  enhanced jigsaw,
  colback=white, colframe=black!60,
  title=Case Study – Multiple Intracranial Lesions in an Adolescent,
  width=\textwidth,
  enlarge top initially by=0mm,
  enlarge bottom finally by=0mm,
  fonttitle=\bfseries,
  fontupper=\normalsize
]

\textbf{Clinical History:} \textit{A previously well 17-year-old boy was admitted to the emergency department with a two-week history of progressive headaches, nausea, and vomiting. This was followed by a single seizure with transient loss of consciousness.
CT demonstrated multiple intracranial lesions with variable enhancement (below; a = precontrast, b = postcontrast). The patient went on to have an MRI scan (below; a = axial T2 images, multiple levels; b = corresponding axial apparent diffusion coefficient [ADC] images; c = corresponding axial T1 postcontrast images).}

\vspace{0.8em}
\begin{tcolorbox}[colback=gray!5, colframe=black!60, title=Phase I – Provided Imaging, breakable]
\begin{center}
\includegraphics[height=2.2cm]{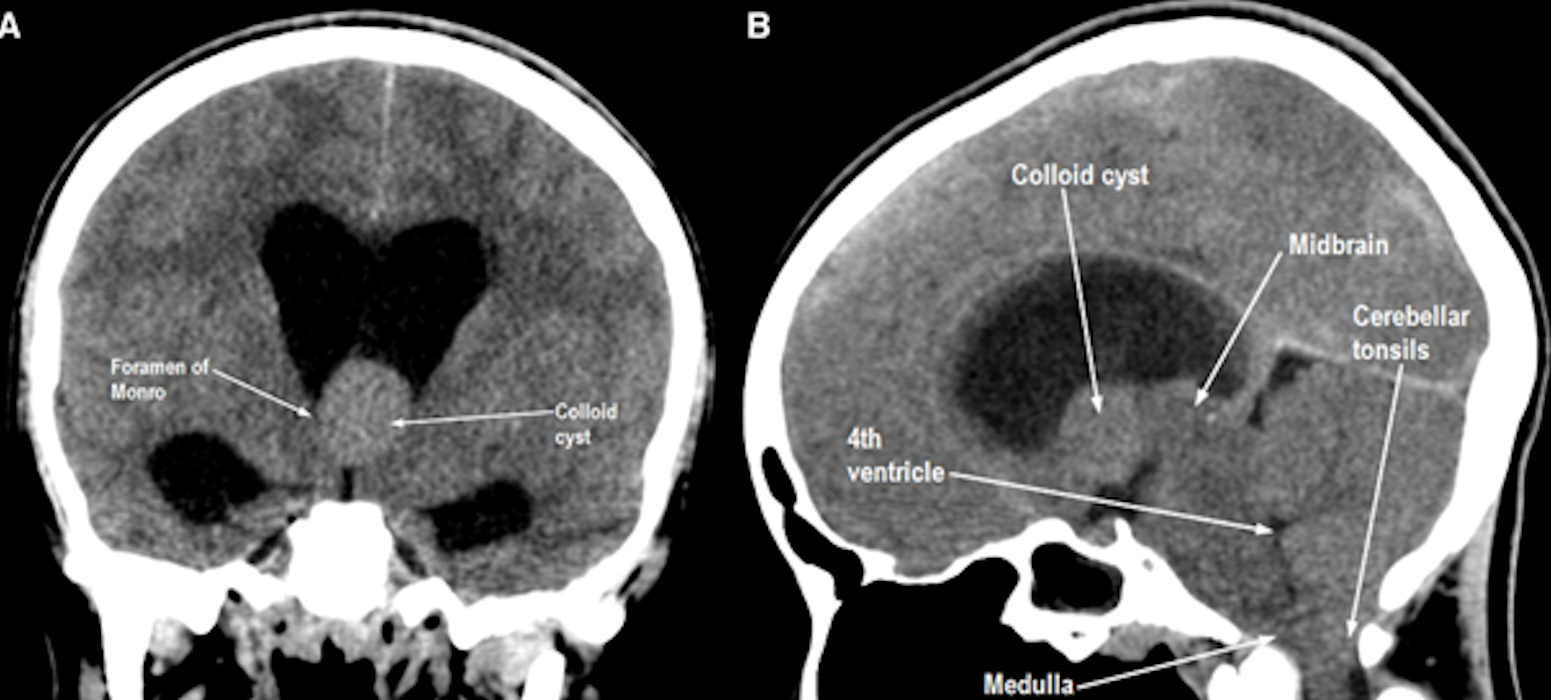}
\hspace{1em}
\includegraphics[height=2.2cm]{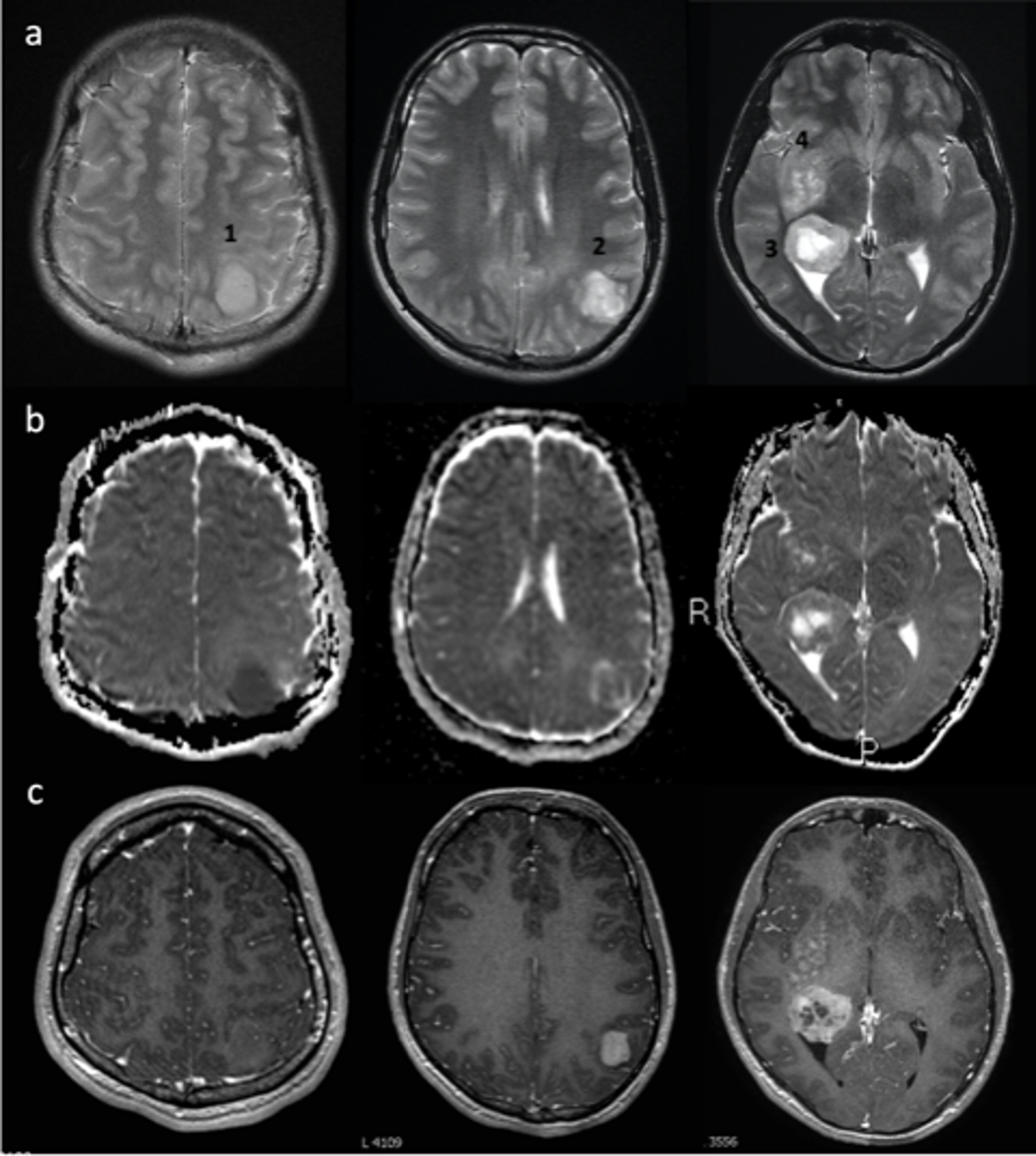}
\end{center}

\vspace{1em}
\textbf{Q1:} What is the likely diagnosis? \\
\textbf{Options:} \\
A. Intracranial metastases \\
B. Germ cell tumor \\
C. Glioblastoma \\
D. Acute disseminated encephalomyelitis (ADEM) \\
E. Lymphoma \\[0.5em]
\textbf{Prediction: A. Intracranial metastases \hfill \textcolor{red}{\textbf{  Incorrect}}} \\
\textbf{Ground Truth: C} \\
\textbf{Explanation:} The patient presents with multiple intracranial lesions that show variable enhancement post-contrast. The imaging characteristics (multiple enhancing lesions with evident mass effect and surrounding edema, as seen in both the CT and MRI scans) are more consistent with intracranial metastases. Other options such as germ cell tumor, glioblastoma, ADEM, and lymphoma typically present with different radiologic patterns, often not as multiple distinct lesions with variable enhancement seen in this case.

\end{tcolorbox}

\end{tcolorbox}
\vspace{0.5em}
\caption{ VQA Case Study – Multiple Intracranial Lesions in an Adolescent}
\label{fig:case-intracranial-lesions}
\end{figure*}

\clearpage

\begin{figure*}[ht]
\begin{tcolorbox}[
  enhanced jigsaw,
  colback=white, colframe=black!60,
  title=Case Study – CTA Assessment of Chronic Abdominal Pain (Round 1),
  width=\textwidth,
  enlarge top initially by=0mm,
  enlarge bottom finally by=0mm,
  fonttitle=\bfseries,
  fontupper=\normalsize
]

\textbf{Clinical History:} \textit{A 19-year-old man presented with several months of intermittent abdominal pain, bloating, and chest pressure. A CTA abdomen pelvis was obtained. Select axial and sagittal images in the arterial and venous phase are shown.}

\vspace{0.8em}
\begin{tcolorbox}[colback=gray!5, colframe=black!60, title=Phase I – CTA Imaging, breakable]
\begin{center}
\includegraphics[height=2.2cm]{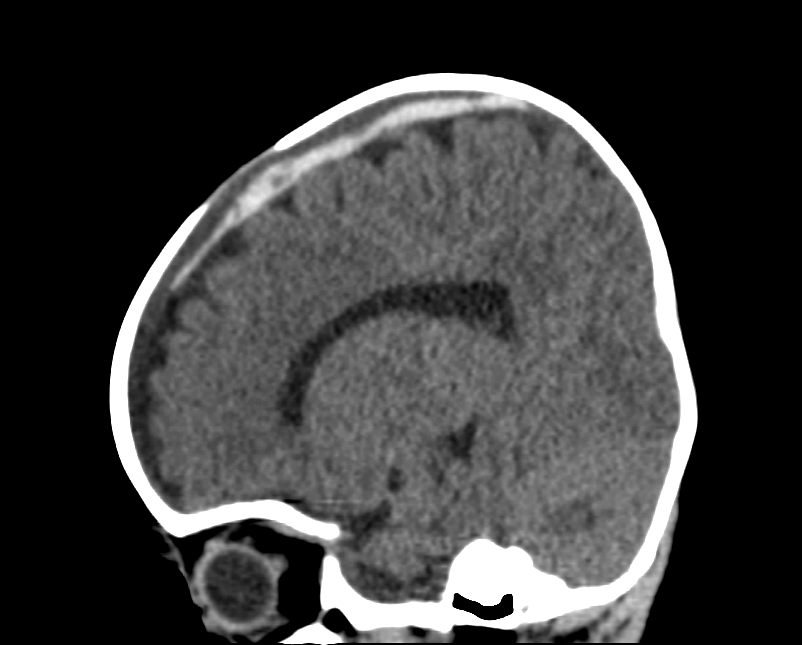}
\hspace{1em}
\includegraphics[height=2.2cm]{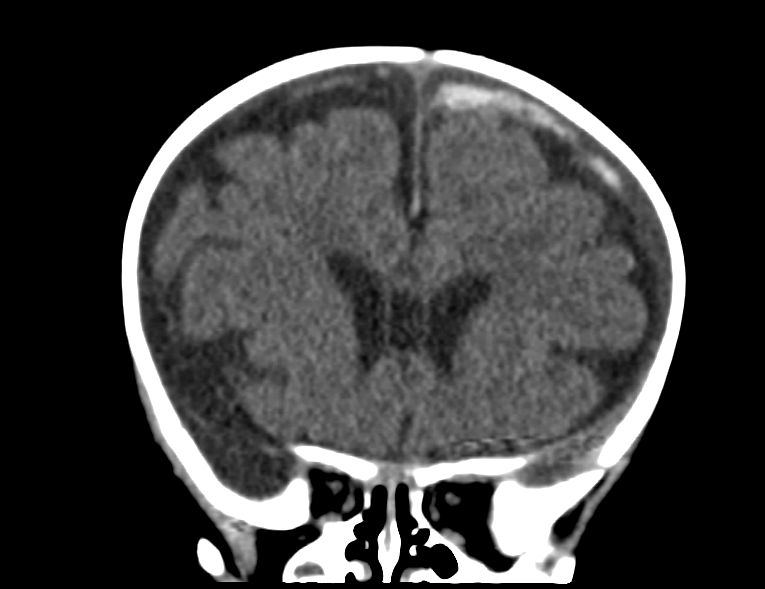}
\hspace{1em}
\includegraphics[height=2.2cm]{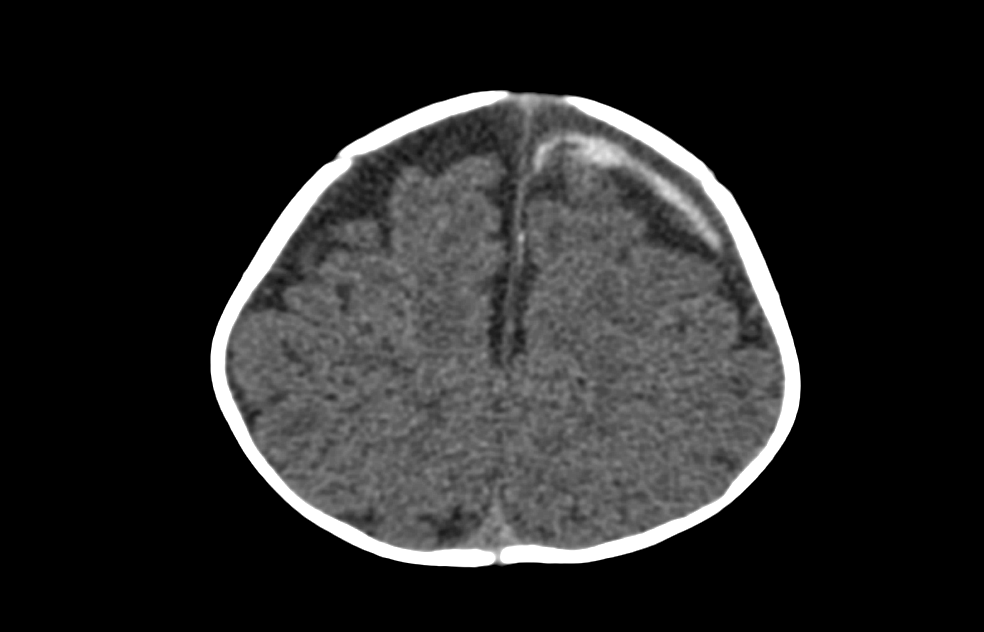}
\end{center}

\vspace{1em}
\textbf{Q1:} Inflammatory fat stranding is present. \\
\textbf{Options:} \\
A. True \\
B. False \\[0.5em]
\textbf{Prediction: B. False \hfill \textcolor{green!50!black}{\textbf{  Correct}}} \\
\textbf{Ground Truth: B} \\
\textbf{Explanation:} Inflammatory fat stranding typically appears as streaky areas of increased attenuation in the fat surrounding organs, indicative of inflammation. The provided images do not show evident fat stranding.

\vspace{1em}
\textbf{Q2:} There is notable atherosclerotic plaque burden resulting in compromised abdominal blood flow. \\
\textbf{Options:} \\
A. True \\
B. False \\[0.5em]
\textbf{Prediction: B. False \hfill \textcolor{green!50!black}{\textbf{  Correct}}} \\
\textbf{Ground Truth: B} \\
\textbf{Explanation:} The images do not demonstrate significant atherosclerotic plaques within the major abdominal arteries that could compromise blood flow. The vessels appear patent without notable narrowing or obstruction.

\end{tcolorbox}

\end{tcolorbox}
\vspace{0.5em}
\caption*{\raggedright \textbf{figure*:} VQA Case Study – CTA Assessment of Chronic Abdominal Pain (Round 1)}
\label{fig:case-abd-pain-r1}
\end{figure*}

\clearpage

\begin{figure*}[ht]
\begin{tcolorbox}[
  enhanced jigsaw,
  colback=white, colframe=black!60,
  title=Case Study – Median Arcuate Ligament Syndrome on CTA (Round 2),
  width=\textwidth,
  enlarge top initially by=0mm,
  enlarge bottom finally by=0mm,
  fonttitle=\bfseries,
  fontupper=\normalsize
]

\textbf{Clinical History:} \textit{A CTA abdomen pelvis was obtained. Select maximum intensity projection (MIP) and 3D volume-rendered images of the celiac axis in the arterial phase during inspiration are shown.}

\vspace{0.8em}
\begin{tcolorbox}[colback=gray!5, colframe=black!60, title=Phase II – MIP and 3D Imaging, breakable]
\begin{center}
\includegraphics[height=2.2cm]{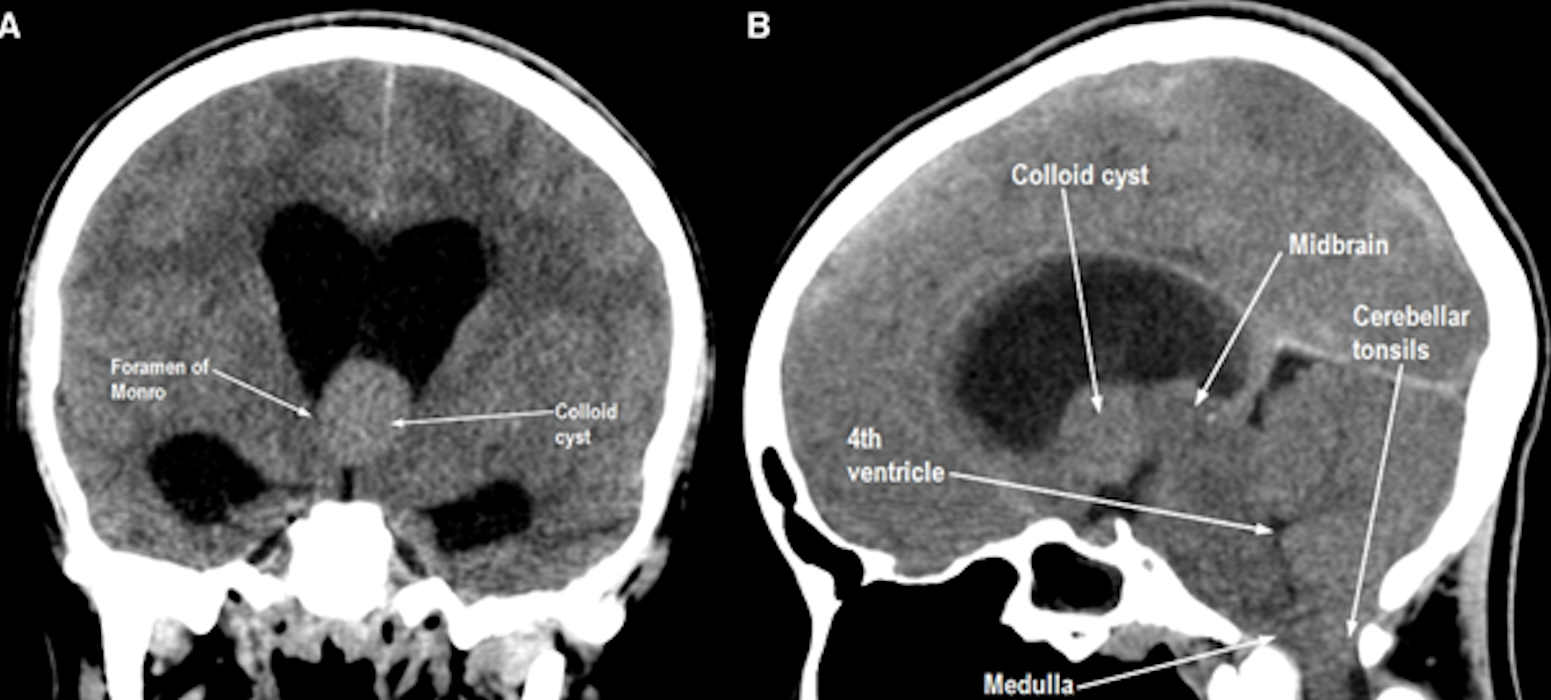}
\hspace{1em}
\includegraphics[height=2.2cm]{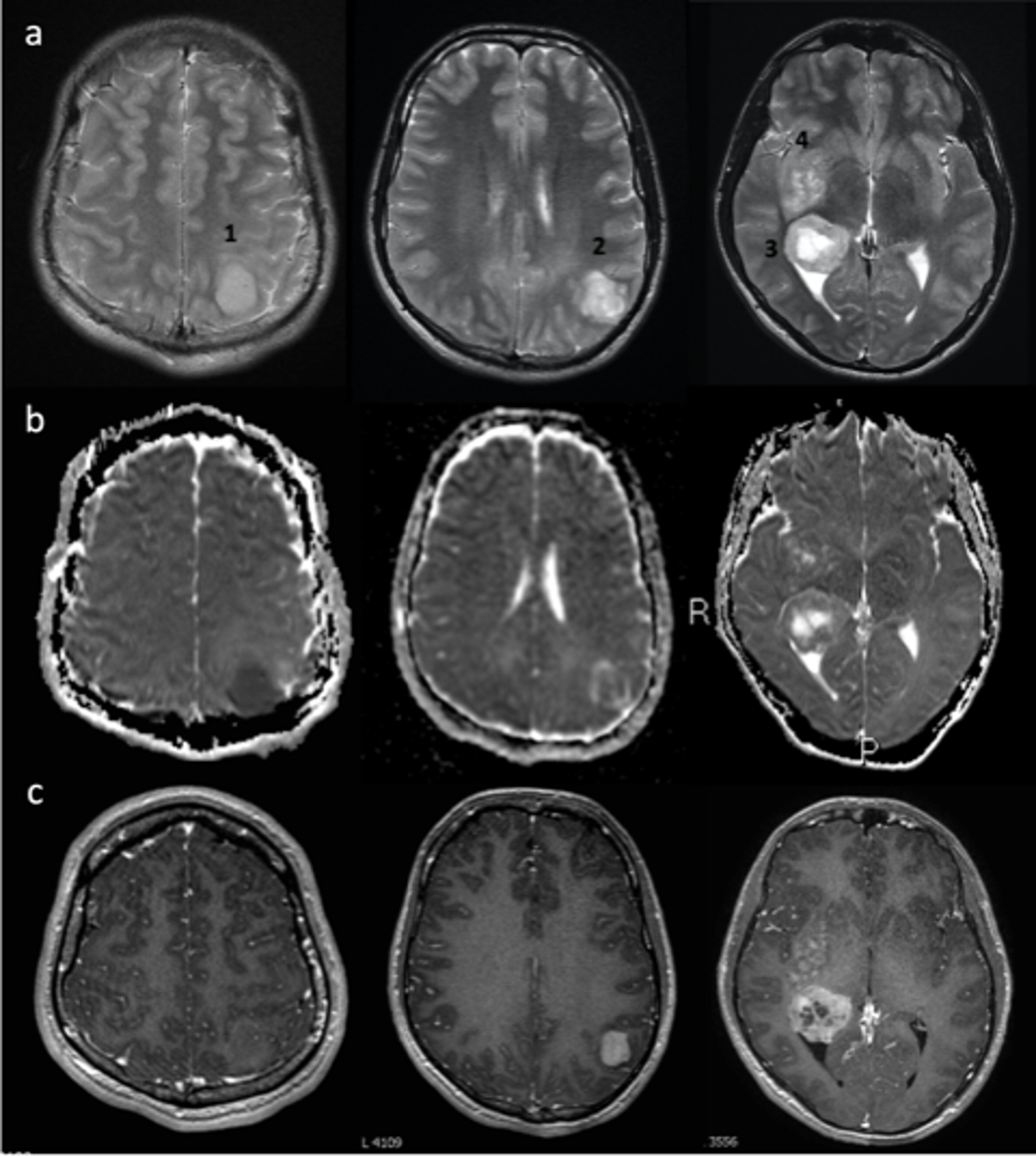}
\end{center}

\vspace{1em}
\textbf{Q1:} Which of the following best characterizes the finding in the images above? \\
\textbf{Options:} \\
A. Compression of the celiac trunk by an external structure \\
B. Atherosclerotic narrowing of the ostium of the celiac artery \\
C. Aneurysm of the celiac artery \\
D. None of the above \\[0.5em]
\textbf{Prediction: A. Compression of the celiac trunk by an external structure \hfill \textcolor{green!50!black}{\textbf{  Correct}}} \\
\textbf{Ground Truth: A} \\
\textbf{Explanation:} The images depict compression of the celiac trunk, which is indicative of median arcuate ligament syndrome. This is commonly caused by external compression from the median arcuate ligament.

\vspace{1em}
\textbf{Q2:} This characteristic finding is usually exaggerated on expiration. \\
\textbf{Options:} \\
A. True \\
B. False \\[0.5em]
\textbf{Prediction: A. True \hfill \textcolor{green!50!black}{\textbf{  Correct}}} \\
\textbf{Ground Truth: A} \\
\textbf{Explanation:} True. The compression of the celiac trunk caused by the median arcuate ligament is usually more prominent on expiration when the diaphragm moves upward, exacerbating the pressure on the celiac artery.

\vspace{1em}
\textbf{Q3:} Which of the following additional imaging features may be seen in patients with median arcuate ligament syndrome? \\
\textbf{Options:} \\
A. Collateralized arterial blood flow \\
B. Post-stenotic dilation \\
C. Absence of associated atherosclerosis \\
D. All of the above \\[0.5em]
\textbf{Prediction: D. All of the above \hfill \textcolor{green!50!black}{\textbf{  Correct}}} \\
\textbf{Ground Truth: D} \\
\textbf{Explanation:} All of the above. Patients with median arcuate ligament syndrome may exhibit collateralized arterial blood flow to compensate for reduced flow, post-stenotic dilation distal to the compression site, and generally do not show atherosclerotic changes associated with this particular syndrome.

\end{tcolorbox}

\end{tcolorbox}
\vspace{0.5em}
\caption*{\raggedright \textbf{figure*:} VQA Case Study – Median Arcuate Ligament Syndrome on CTA (Round 2)}
\label{fig:case-mals-r2}
\end{figure*}

\clearpage

\begin{figure*}[ht]
\begin{tcolorbox}[
  enhanced jigsaw,
  colback=white, colframe=black!60,
  title=Case Study – Post-Ureteroscopy Flank Pain and Retroperitoneal Hematoma (Round 1),
  width=\textwidth,
  enlarge top initially by=0mm,
  enlarge bottom finally by=0mm,
  fonttitle=\bfseries,
  fontupper=\normalsize
]

\textbf{Clinical History:} \textit{A 19-year-old man presented with left flank pain seven days after ureteroscopy and urinary calculus removal. Unenhanced CT images are shown below.}

\vspace{0.8em}
\begin{tcolorbox}[colback=gray!5, colframe=black!60, title=Phase I – CT Imaging, breakable]
\begin{center}
\includegraphics[height=2.2cm]{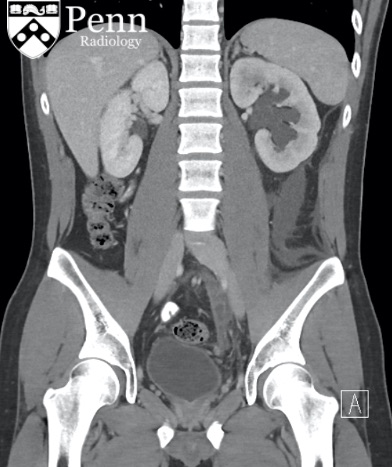}
\hspace{1em}
\includegraphics[height=2.2cm]{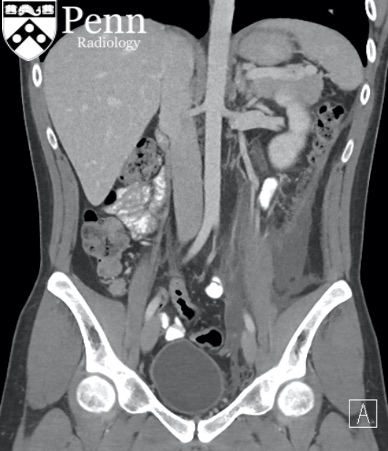}
\hspace{1em}
\includegraphics[height=2.2cm]{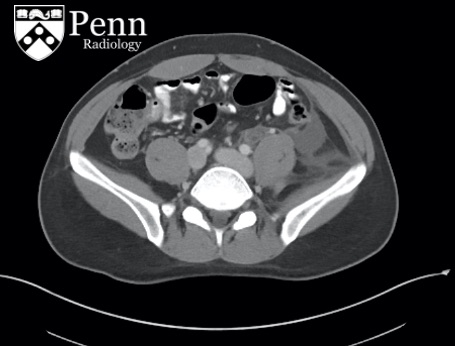}
\end{center}

\vspace{1em}
\textbf{Q1:} What is the likely underlying pathophysiology of these findings? \\
\textbf{Options:} \\
A. Iatrogenic injury \\
B. Infection \\
C. Retained ureteric calculus \\[0.5em]
\textbf{Prediction: A. Iatrogenic injury \hfill \textcolor{green!50!black}{\textbf{  Correct}}} \\
\textbf{Ground Truth: A} \\
\textbf{Explanation:} The CT images show a large fluid collection around the left kidney with high density characteristics, suggesting a hematoma. This recent history of ureteroscopy and urinary calculus removal points towards an iatrogenic injury as the likely underlying pathophysiology.

\vspace{1em}
\textbf{Q2:} What Hounsfield unit (HU) is most likely for the abnormality in the left retroperitoneum? \\
\textbf{Options:} \\
A. 0-10 HU \\
B. 20-30 HU \\
C. 40-50 HU \\
D. > 80 HU \\[0.5em]
\textbf{Prediction: D. > 80 HU \hfill \textcolor{red}{\textbf{  Incorrect}}} \\
\textbf{Ground Truth: A} \\
\textbf{Explanation:} Although clotted blood can sometimes measure higher, in many retroperitoneal hematomas, the HU may fall closer to fluid densities (e.g., 0–10 HU) early on, especially when unenhanced. CT attenuation values are important in determining hemorrhagic vs. serous collections.

\vspace{1em}
\textbf{Q3:} What is the next best management step? \\
\textbf{Options:} \\
A. Excretory phase CT \\
B. Follow-up abdominal radiograph \\
C. Reassurance and analgesia \\
D. Percutaneous aspiration \\[0.5em]
\textbf{Prediction: D. Percutaneous aspiration \hfill \textcolor{red}{\textbf{  Incorrect}}} \\
\textbf{Ground Truth: A} \\
\textbf{Explanation:} An excretory phase CT is important to assess for ongoing urine leak or injury to the collecting system following ureteral procedures. Aspiration may be premature without knowing the nature of the fluid.

\end{tcolorbox}

\end{tcolorbox}
\vspace{0.5em}
\caption*{\raggedright \textbf{figure*:} VQA Case Study – Post-Ureteroscopy Flank Pain and Retroperitoneal Hematoma (Round 1)}
\label{fig:case-ureteroscopy-r1}
\end{figure*}

\clearpage

\begin{figure*}[ht]
\begin{tcolorbox}[
  enhanced jigsaw,
  colback=white, colframe=black!60,
  title=Case Study – Retrograde Pyelogram Following Iatrogenic Ureteral Injury (Round 2),
  width=\textwidth,
  enlarge top initially by=0mm,
  enlarge bottom finally by=0mm,
  fonttitle=\bfseries,
  fontupper=\normalsize
]

\textbf{Clinical History:} \textit{A follow-up radiograph was performed 30 minutes after the initial CT. Based on the CT and radiograph findings, the patient was brought for cystoscopy and retrograde pyelogram. A retrograde pyelogram image is shown below.}

\vspace{0.8em}
\begin{tcolorbox}[colback=gray!5, colframe=black!60, title=Phase II – Pyelogram Imaging, breakable]
\begin{center}
\includegraphics[height=2.2cm]{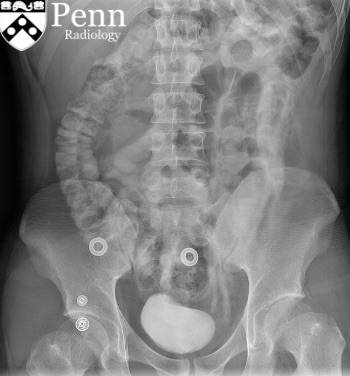}
\hspace{1em}
\includegraphics[height=2.2cm]{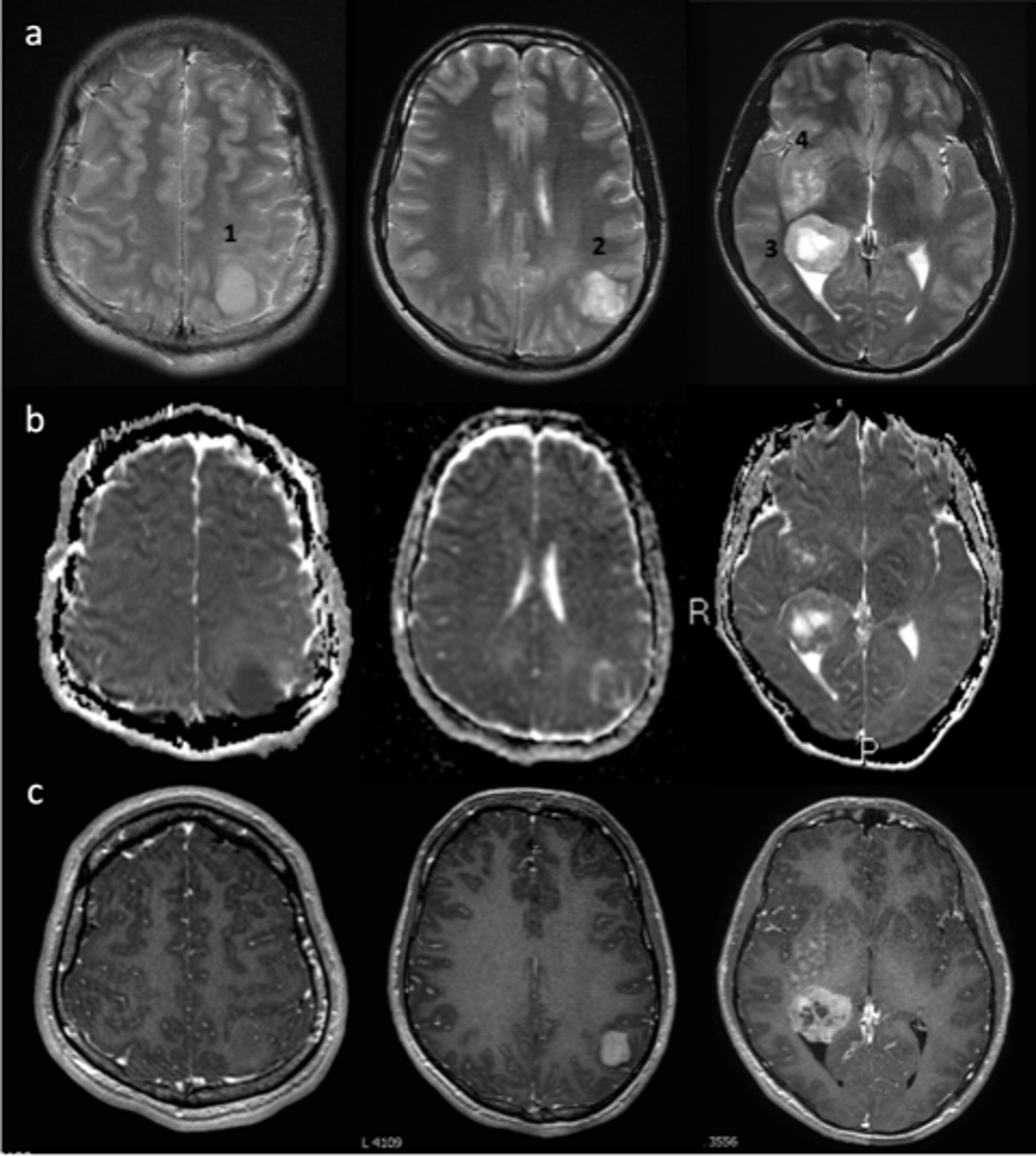}
\end{center}

\vspace{1em}
\textbf{Q1:} The delayed presentation for this patient is typical for iatrogenic ureteral injuries. \\
\textbf{Options:} \\
A. TRUE \\
B. FALSE \\[0.5em]
\textbf{Prediction: A. TRUE \hfill \textcolor{green!50!black}{\textbf{  Correct}}} \\
\textbf{Ground Truth: A} \\
\textbf{Explanation:} Iatrogenic ureteral injuries often present with delayed symptoms after the initial procedure, commonly several days to a week later, as seen in this patient who presented with left flank pain seven days post-ureteroscopy.

\vspace{1em}
\textbf{Q2:} This ureteral injury will require reimplantation. \\
\textbf{Options:} \\
A. True \\
B. FALSE \\[0.5em]
\textbf{Prediction: B. FALSE \hfill \textcolor{green!50!black}{\textbf{  Correct}}} \\
\textbf{Ground Truth: B} \\
\textbf{Explanation:} The retrograde pyelogram shows extravasation of contrast suggestive of a potential ureteral injury. However, not all ureteral injuries require reimplantation. Many can be managed with stenting depending on location and severity.

\end{tcolorbox}

\end{tcolorbox}
\vspace{0.5em}
\caption*{\raggedright \textbf{figure*:} VQA Case Study – Retrograde Pyelogram Following Iatrogenic Ureteral Injury (Round 2)}
\label{fig:case-ureteroscopy-r2}
\end{figure*}

\clearpage

% Round 1
\begin{figure*}[ht]
\begin{tcolorbox}[
  enhanced jigsaw,
  colback=white, colframe=black!60,
  title=Case Study – Breast Pain with Cordlike Mass (Round 1),
  width=\textwidth,
  fonttitle=\bfseries,
  fontupper=\normalsize
]

\textbf{Clinical History:} \textit{A 38-year-old woman presented with unilateral breast pain she had experienced for a few weeks, which was associated with a palpable cordlike mass and erythema. A bilateral screening mammogram with tomosynthesis was obtained. The right mediolateral oblique (MLO) view of the mammogram is shown below with the area of concern marked with metallic BB. A zoomed-in image centered on the area of concern is shown next.}

\vspace{0.8em}
\begin{center}
\includegraphics[height=4cm]{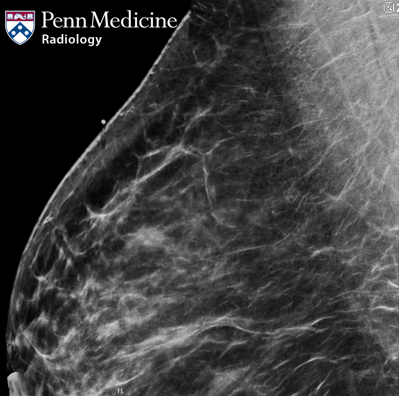}
\hspace{1em}
\includegraphics[height=4cm]{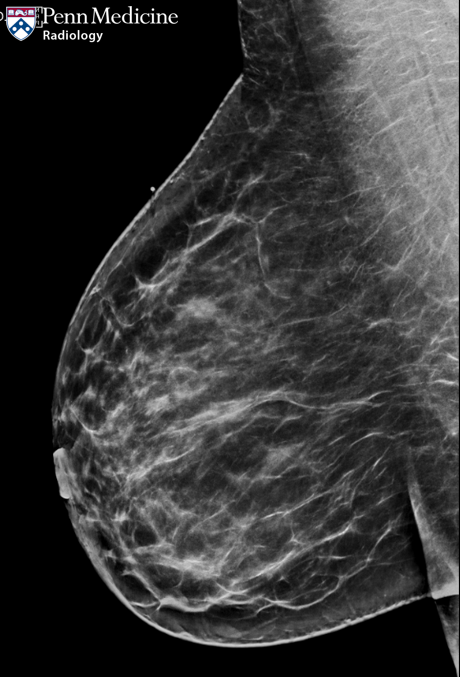}
\end{center}

\vspace{1em}
\textbf{Q1:} The density of this breast would be categorized as category A, almost entirely fat. \\
\textbf{Options:} A. TRUE \quad B. FALSE \\[0.5em]
\textbf{Prediction: B. FALSE \hfill \textcolor{green!50!black}{\textbf{  Correct}}} \\
\textbf{Ground Truth: B} \\
\textbf{Explanation:} The breast tissue shows a mix of fibroglandular and fatty tissue, thus not meeting BI-RADS category A criteria (which is almost entirely fat). White regions on the mammogram indicate glandular density.

\end{tcolorbox}
\caption{VQA Case Study – Breast Pain (Round 1)}
\label{fig:case-breast-round1}
\end{figure*}

\clearpage

% Round 2
\begin{figure*}[ht]
\begin{tcolorbox}[
  enhanced jigsaw,
  colback=white, colframe=black!60,
  title=Case Study – Breast Pain with Cordlike Mass (Round 2),
  width=\textwidth,
  fonttitle=\bfseries,
  fontupper=\normalsize
]

\textbf{Imaging:} \textit{Tomosynthesis of the bilateral breasts was obtained. The right MLO view is shown.}

\vspace{0.8em}
\begin{center}
\includegraphics[height=4.5cm]{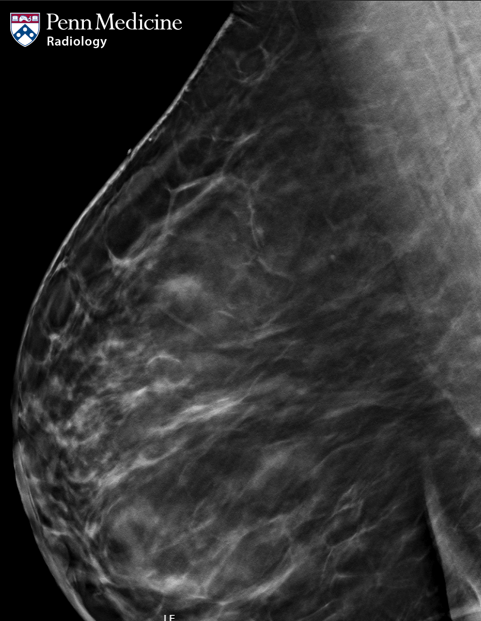}
\end{center}

\vspace{1em}
\textbf{Q1:} In the workup of a palpable lump without obvious mammographic findings, what is the appropriate next step? \\
\textbf{Options:} A. Ultrasound \quad B. Fine-needle aspiration \quad C. MRI \quad D. Repeat imaging in two weeks \\[0.5em]
\textbf{Prediction: A. Ultrasound \hfill \textcolor{green!50!black}{\textbf{  Correct}}} \\
\textbf{Ground Truth: A} \\
\textbf{Explanation:} Ultrasound is the preferred next step for evaluating a palpable lump when mammogram findings are inconclusive, particularly to differentiate solid vs cystic lesions.

\end{tcolorbox}
\caption{VQA Case Study – Breast Pain (Round 2)}
\label{fig:case-breast-round2}
\end{figure*}

\clearpage

% Round 3
\begin{figure*}[ht]
\begin{tcolorbox}[
  enhanced jigsaw,
  colback=white, colframe=black!60,
  title=Case Study – Breast Pain with Cordlike Mass (Round 3),
  width=\textwidth,
  fonttitle=\bfseries,
  fontupper=\normalsize
]

\textbf{Imaging:} \textit{Ultrasound images of the right breast were obtained focusing on the area of concern. Compression views and color Doppler were applied. The first two images depict the structure in longitudinal view, and the third in transverse view.}

\vspace{0.8em}
\begin{center}
\includegraphics[height=3.5cm]{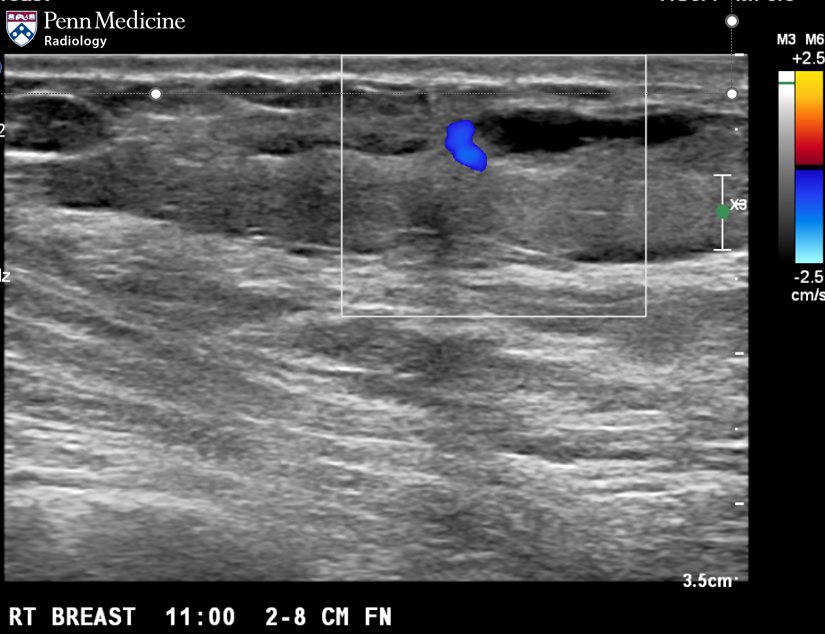}
\hspace{0.5em}
\includegraphics[height=3.5cm]{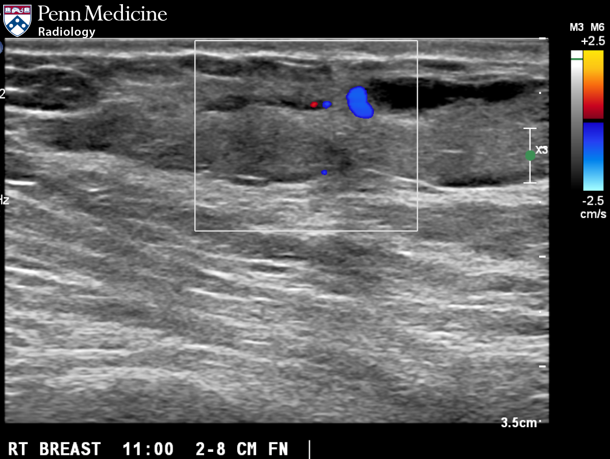}
\hspace{0.5em}
\includegraphics[height=3.5cm]{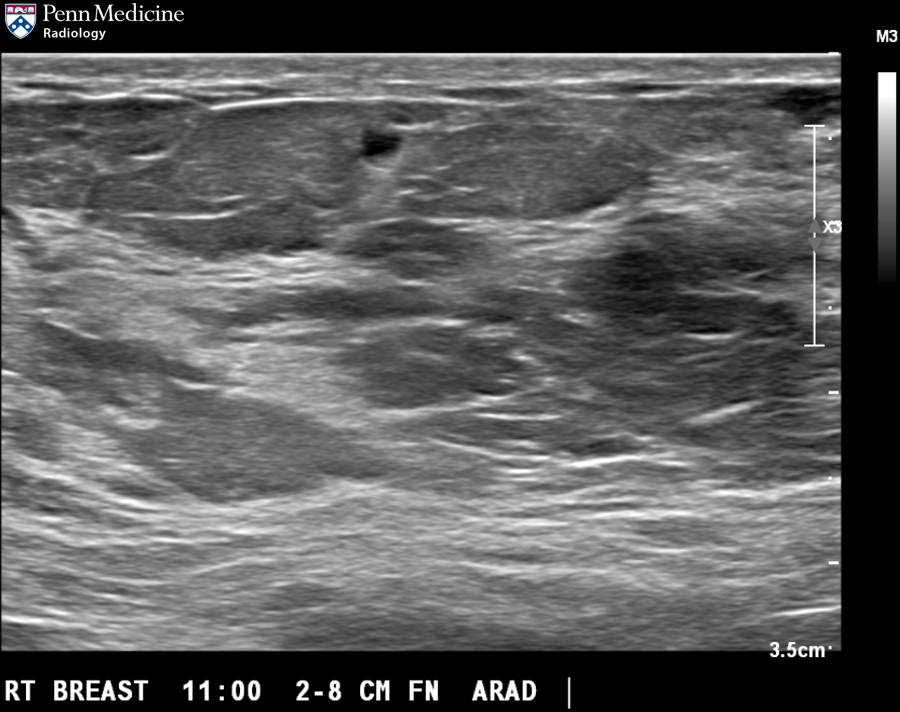}
\end{center}

\vspace{1em}
\textbf{Q1:} What do the ultrasound images depict? \\
\textbf{Options:} A. Localized fluid collection \quad B. Intraductal mass \quad C. Shadowing mass \quad D. Inflamed fat lobules \quad E. Hypoechoic tubular structure without vascular flow \\[0.5em]
\textbf{Prediction: E. Hypoechoic tubular structure without vascular flow \hfill \textcolor{green!50!black}{\textbf{  Correct}}} \\
\textbf{Ground Truth: E} \\
\textbf{Explanation:} This is consistent with a thrombosed vessel, often appearing as a hypoechoic non-compressible tubular structure without color Doppler flow.

\vspace{1em}
\textbf{Q2:} The clinical presentation of a thrombosed vessel can include all of the following EXCEPT: \\
\textbf{Options:} A. Bloody nipple discharge \quad B. Unilateral pain \quad C. Cordlike mass \quad D. Skin erythema \\[0.5em]
\textbf{Prediction: A. Bloody nipple discharge \hfill \textcolor{green!50!black}{\textbf{  Correct}}} \\
\textbf{Ground Truth: A} \\
\textbf{Explanation:} Bloody nipple discharge is not characteristic of thrombosed vessels. The other symptoms are common findings in superficial thrombophlebitis of the breast.

\end{tcolorbox}
\caption{VQA Case Study – Breast Pain (Round 3)}
\label{fig:case-breast-round3}
\end{figure*}

\clearpage
\begin{figure*}[ht]
\begin{tcolorbox}[
  enhanced jigsaw,
  colback=white, colframe=black!60,
  title=Case Study – Chronic Shoulder Pain in 39-Year-Old Woman (Round 1),
  width=\textwidth,
  fonttitle=\bfseries,
  fontupper=\normalsize,
  breakable
]

\textbf{Clinical History:} \textit{A 39-year-old woman presented with chronic atraumatic right shoulder pain, which persists at exertion and at rest. She denies any inciting event that triggered the pain. Physical exam localizes the pain to the superolateral shoulder, with preservation of range of motion. The patient underwent radiographs for initial evaluation.}

\vspace{0.8em}
\begin{center}
\includegraphics[height=2.8cm]{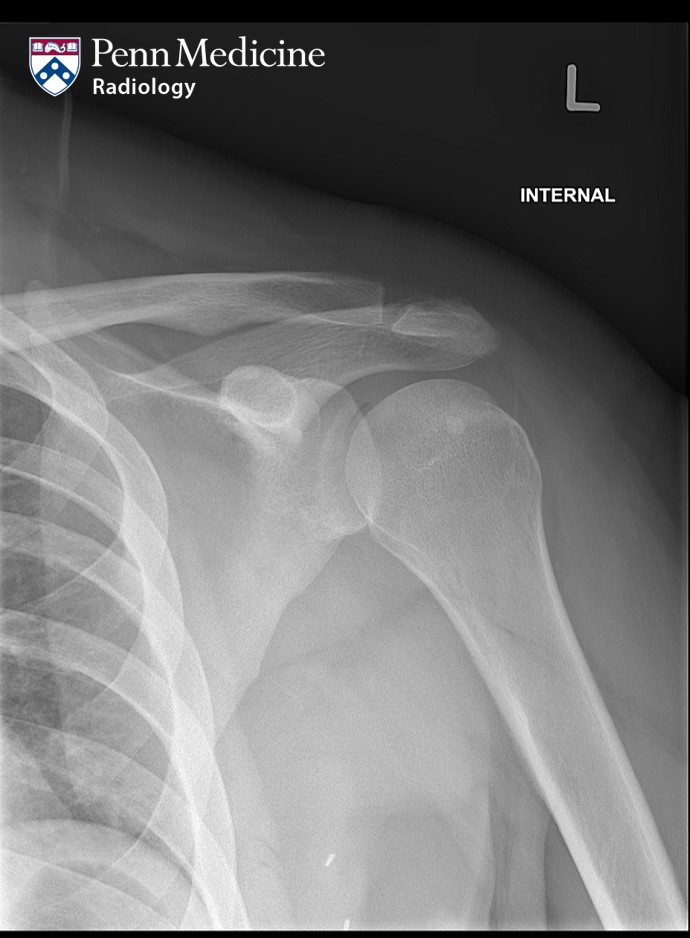}
\hspace{0.5em}
\includegraphics[height=2.8cm]{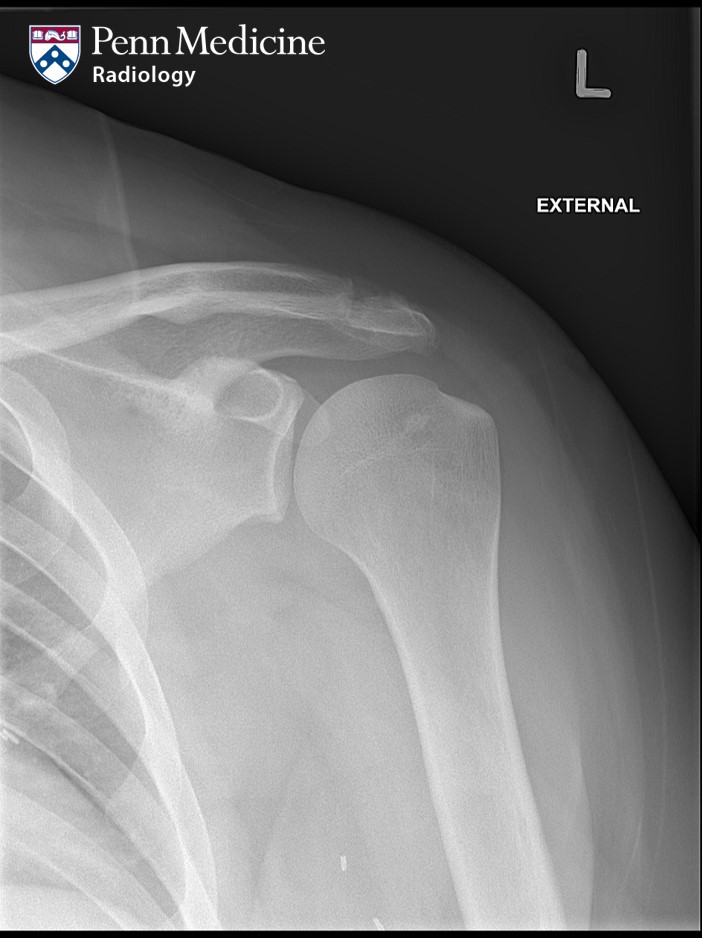}
\hspace{0.5em}
\includegraphics[height=2.8cm]{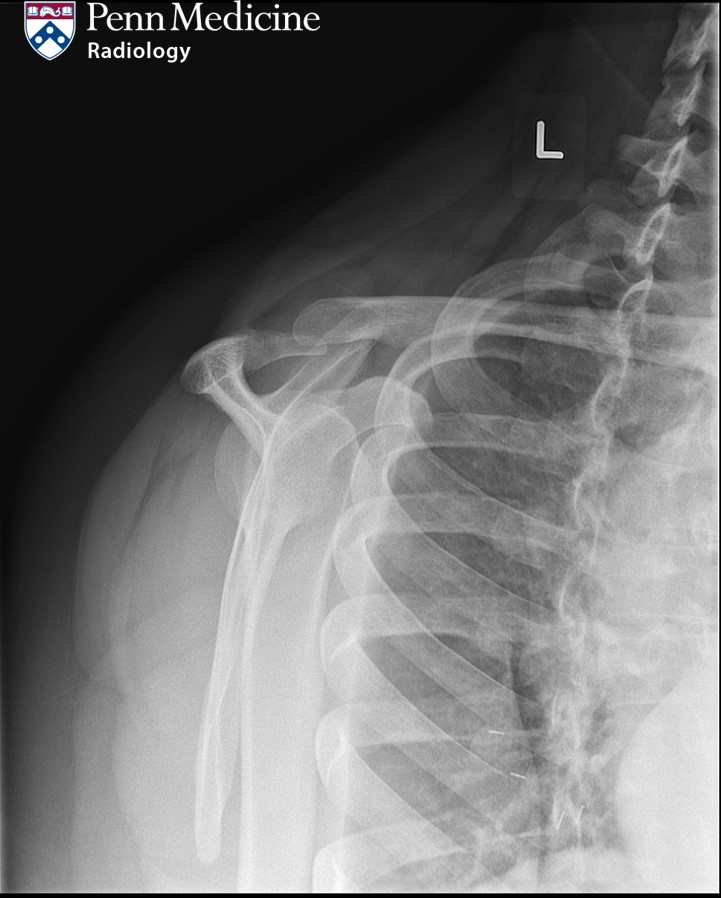}
\hspace{0.5em}
\includegraphics[height=2.8cm]{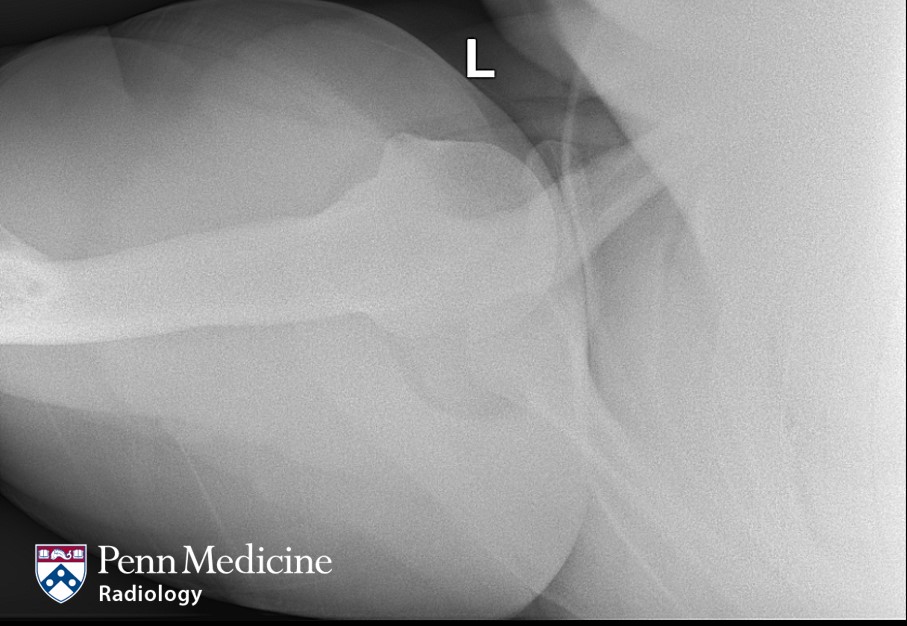}
\end{center}

\vspace{0.8em}
\textbf{Q1:} What is the main finding? \\
\textbf{Options:} \\
A. Fracture of the humeral neck \\
B. Widening of the acromioclavicular joint \\
C. Anterior shoulder dislocation \\
D. Severe osteoarthritis of the glenohumeral joint \\
E. No radiographic abnormality is visible \\[0.5em]
\textbf{Prediction: E. No radiographic abnormality is visible \hfill \textcolor{green!50!black}{\textbf{  Correct}}} \\
\textbf{Ground Truth: E} \\
\textbf{Explanation:} The provided X-ray images do not demonstrate fracture, dislocation, joint space widening, or arthritic changes. The acromioclavicular joint and humeral alignment are intact.

\end{tcolorbox}
\caption{VQA Case Study – Shoulder Pain (Round 1)}
\label{fig:case-shoulder-1}
\end{figure*}

\clearpage

\begin{figure*}[ht]
\begin{tcolorbox}[
  enhanced jigsaw,
  colback=white, colframe=black!60,
  title=Case Study – Shoulder MRI Findings (Round 2),
  width=\textwidth,
  fonttitle=\bfseries,
  fontupper=\normalsize,
  breakable
]

\textbf{Clinical History:} \textit{The patient subsequently underwent MR of the shoulder for further evaluation. Select axial proton density/fluid-sensitive images are shown.}

\vspace{0.8em}
\begin{center}
\includegraphics[height=3.5cm]{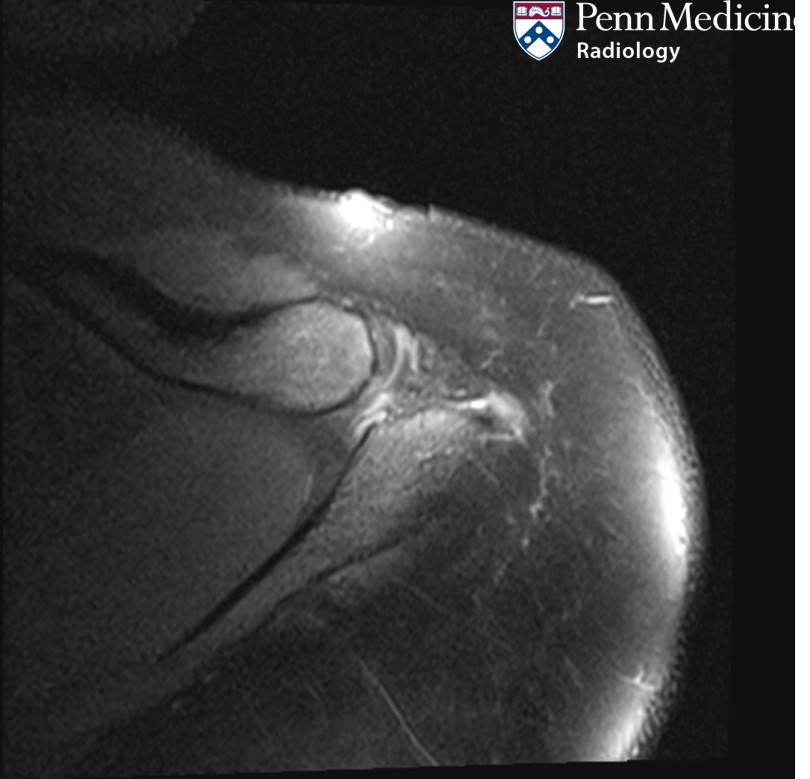}
\hspace{0.5em}
\includegraphics[height=3.5cm]{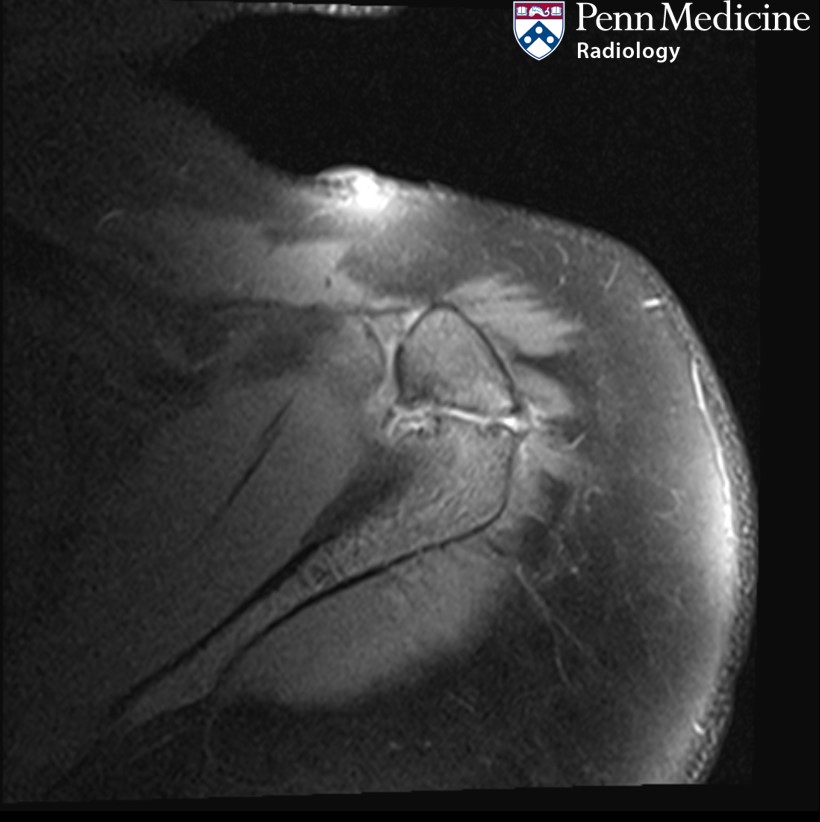}
\end{center}

\vspace{0.8em}
\textbf{Q1:} There is an os acromiale. \\
\textbf{Options:} \\
A. TRUE \\
B. FALSE \\[0.5em]
\textbf{Prediction: A. TRUE \hfill \textcolor{green!50!black}{\textbf{  Correct}}} \\
\textbf{Ground Truth: A} \\
\textbf{Explanation:} A non-fused segment of the acromion with well-corticated margins is indicative of os acromiale.

\vspace{1em}
\textbf{Q2:} What subtype does this patient have? \\
\textbf{Options:} \\
A. Pre-acromion \\
B. Meso-acromion \\
C. Meta-acromion \\
D. Unable to tell \\[0.5em]
\textbf{Prediction: D. Unable to tell \hfill \textcolor{red}{\textbf{  Incorrect}}} \\
\textbf{Ground Truth: B} \\
\textbf{Explanation:} Based on the location of the unfused fragment relative to the acromioclavicular joint and scapular spine, this is consistent with meso-acromion.

\vspace{1em}
\textbf{Q3:} There are MR findings to indicate acute inflammation. \\
\textbf{Options:} \\
A. TRUE \\
B. FALSE \\[0.5em]
\textbf{Prediction: B. FALSE \hfill \textcolor{red}{\textbf{  Incorrect}}} \\
\textbf{Ground Truth: A} \\
\textbf{Explanation:} The MR images show signal abnormalities consistent with edema and inflammation, consistent with acute inflammatory changes.

\vspace{1em}
\textbf{Q4:} Most people with this anatomic variant are symptomatic. \\
\textbf{Options:} \\
A. TRUE \\
B. FALSE \\[0.5em]
\textbf{Prediction: B. FALSE \hfill \textcolor{green!50!black}{\textbf{  Correct}}} \\
\textbf{Ground Truth: B} \\
\textbf{Explanation:} Os acromiale is often asymptomatic and discovered incidentally on imaging studies.

\end{tcolorbox}
\caption{VQA Case Study – Shoulder MRI Findings (Round 2)}
\label{fig:case-shoulder-2}
\end{figure*}

\end{document}